\documentclass{article}

\usepackage[final]{nips_2017}

\usepackage[utf8]{inputenc} 
\usepackage[T1]{fontenc}    
\usepackage{hyperref}       
\usepackage{booktabs}       
\usepackage{amsfonts}       
\usepackage{nicefrac}       
\usepackage{microtype}      

\usepackage{graphicx,dblfloatfix} 
\usepackage{subcaption}
\usepackage{times}
\usepackage{url}
\usepackage{color}
\usepackage{amsmath,bm}
\usepackage{amssymb}
\usepackage{multirow}
\usepackage{comment}
\usepackage{multicol}
\usepackage{algorithm}
\usepackage{algorithmic}
\usepackage{xspace}
\usepackage[algo2e, noend, noline, ruled]{algorithm2e}
\usepackage{wrapfig}

\providecommand{\DontPrintSemicolon}{\dontprintsemicolon}
\DontPrintSemicolon


\newif\ifspacetight
\spacetighttrue

\ifspacetight

\definecolor{grey}{rgb}{0.7,0.7,0.7}

\makeatletter
\let\@listiold\@listi
\let\@listiiold\@listii
\let\@listiiiold\@listiii
\let\@listivold\@listiv
\let\@listvold\@listv

\makeatother
\fi

\definecolor{darkgreen}{RGB}{0,125,0}
\definecolor{orange}{RGB}{255,124,84}
\newcounter{mlNoteCounter}

\newcommand{\argmin}{\operatornamewithlimits{argmin}}

\newcommand{\BR}{\mbox{BR}}
\newcommand{\defword}[1]{\textbf{\boldmath{#1}}}
\newcommand{\ie}{{\it i.e.}~}
\newcommand{\eg}{{\it e.g.}~}

\def\ie{{\em i.e.,}\xspace}

\title{A Unified Game-Theoretic Approach to\\Multiagent Reinforcement Learning}

\author{
  Marc Lanctot\\
  DeepMind\\
  \texttt{lanctot@}\\
  \And
  Vinicius Zambaldi\\
  DeepMind\\
  \texttt{vzambaldi@}\\
  \And
  Audr\={u}nas Gruslys\\
  DeepMind\\
  \texttt{audrunas@}\\
  \And
  Angeliki Lazaridou\\
  DeepMind\\
  \texttt{angeliki@}\\
  \And
  Karl Tuyls\\
  DeepMind\\
  \texttt{karltuyls@}\\
  \And
  Julien P\'{e}rolat\\
  DeepMind\\
  \texttt{perolat@}\\
  \And
  David Silver\\
  DeepMind\\
  \texttt{davidsilver@}\\
  \And
  Thore Graepel\\
  DeepMind\\
  \texttt{thore@}\\
  \AND
  \normalfont \texttt{...@google.com}\\
}

\begin{document}

\maketitle

\begin{abstract}
To achieve general intelligence, agents must learn how to interact with
others in a shared environment: this is the challenge of multiagent
reinforcement learning (MARL). The simplest form is
{\it independent reinforcement learning} (InRL), where each agent
treats its experience as part of its (non-stationary) environment.
In this paper, we first observe that policies learned using InRL
can overfit to the other agents' policies during training, failing to sufficiently
generalize during execution. We introduce a new metric,
{\it joint-policy correlation}, to quantify this effect.
We describe an algorithm for general MARL, based on approximate
best responses to mixtures of policies generated using deep reinforcement learning, and
empirical game-theoretic analysis to compute meta-strategies for policy selection.
The algorithm generalizes previous ones such as InRL,
iterated best response, double oracle, and fictitious play.
Then, we present a scalable implementation which reduces the memory requirement using
decoupled meta-solvers.
Finally, we demonstrate the generality of the resulting policies in two partially
observable settings: gridworld coordination games and poker.
\end{abstract}

\section{Introduction}



Deep reinforcement learning combines deep learning~\cite{LeCun15} with reinforcement
learning~\cite{Sutton98,Littman15} to compute a policy used to drive
decision-making~\cite{Mnih15DQN,mnih2016asynchronous}.
Traditionally, a single agent interacts with its environment
repeatedly, iteratively improving its policy by learning from its observations.
Inspired by recent success in Deep RL, we are now seeing a renewed interest in {\it multiagent} reinforcement learning
(MARL)~\cite{ShohamPG07,Busoniu08Comprehensive,TuylsW12}. In MARL, several agents interact and learn in an environment
simultaneously, either competitively such as in Go~\cite{Silver16AlphaGo} and
Poker~\cite{Heinrich15FSP,Yakovenko16,Moravcik17DeepStack},
cooperatively such as when learning to communicate~\cite{Foerster16,Sukhbaatar16,Hausknecht16},
or some mix of the two~\cite{Leibo17SSD,Tampuu17Pong,ICLR16-hausknecht}.

The simplest form of MARL is {\it independent RL} (InRL), where each learner is oblivious to the
other agents and simply treats all the interaction as part of its (``localized'') environment.
Aside from the problem that these local environments are non-stationary and
non-Markovian~\cite{Laurent11} resulting in a loss of convergence guarantees for many algorithms,
the policies found can overfit to the other agents' policies and hence not generalize well.
There has been relatively little work done in RL community on overfitting to the
environment~\cite{Whiteson11,Marivate15}, but we argue that this is particularly important in
multiagent settings where one must react dynamically based on the observed behavior of others.
Classical techniques collect or approximate extra information such as the joint
values~\cite{Littman94markovgames,Claus98Dynamics,Greenwald03CEQ,Lauer04}, use adaptive
learning rates~\cite{02aij-wolf}, adjust the frequencies of updates \cite{KaisersT10,PanaitTL08}, or
dynamically respond to the other agents actions online~\cite{Littman01FFQ,KW16}.
However, with the notable exceptions of very recent work~\cite{Foerster17,Omidshafiei17},
they have focused on (repeated) matrix games and/or the fully-observable case.

There have been several proposals for treating partial observability in the multiagent setting.
When the model is fully known and the setting is strictly adversarial with two players,
there are policy iteration methods based on regret minimization that scale very well when using
domain-specific abstractions~\cite{Gilpin09,brown2015hierarchical,2013aamas-kmeans-abstraction,Johanson16PhD},
which was a major component of the expert no-limit poker AI Libratus~\cite{BrownS17};
recently these methods were combined
with deep learning to create an expert no-limit poker AI called DeepStack~\cite{Moravcik17DeepStack}.
There is a significant amount of work that deals with the case of decentralized
cooperative problems~\cite{Nair04,OliehoekAmato16book}, and in the general setting
by extending the notion of belief states and Bayesian updating from POMDPs~\cite{IPOMDPs}.
These models are quite expressive, and the resulting algorithms are fairly complex.
In practice, researchers often resort to approximate forms, by sampling or exploiting structure,
to ensure good performance due to intractability~\cite{Hoang13,Amato15AAAI,fansaamas08}.

In this paper, we introduce a new metric for quantifying the correlation effects
of policies learned by independent learners, and demonstrate the severity of the overfitting problem.
These coordination problems have been well-studied in the fully-observable cooperative case~\cite{Matignon12Independent}:
we observe similar problems in a partially-observed mixed cooperative/competitive setting and, and we show that
the severity increases as the environment becomes more partially-observed.
We propose a new algorithm based on economic reasoning~\cite{Parkes267}, 
which uses (i) deep reinforcement learning
to compute best responses to a distribution over policies, and (ii) empirical game-theoretic analysis to
compute new meta-strategy distributions.
As is common in the MARL setting, we assume centralized training for decentralized execution: policies are
represented as separate neural networks and there is no sharing of gradients nor architectures among agents.
The basic form uses a centralized payoff table, which is removed in the distributed, decentralized form
that requires less space.


\section{Background and Related Work}


In this section, we start with basic building blocks necessary to describe the algorithm.
We interleave this with the most relevant previous work for our setting. Several components
of the general idea have been (re)discovered many times across different research communities,
each with slightly different but similar motivations. One aim here is therefore to unify the
algorithms and terminology.

A \defword{normal-form game} is a tuple $(\Pi, U, n)$ where $n$ is the number of players,
$\Pi = (\Pi_1, \cdots, \Pi_n)$ is the set of policies (or strategies, one for each player
$i \in [[n]]$, where $[[n]] = \{1, \cdots, n\}$),
and $U : \Pi \rightarrow \Re^n$ is a payoff table of utilities for each joint policy played by
all players.
{\bf Extensive-form games} extend these formalisms to the multistep sequential case (\eg poker).

Players try to maximize their own expected utility.
Each player does this by choosing a policy from $\Pi_i$, or by sampling
from a mixture (distribution) over them $\sigma_i \in \Delta(\Pi_i)$.
In this multiagent setting, the quality of $\sigma_i$ depends on other players'
strategies, and so it cannot be found nor assessed independently.
Every finite extensive-form game has an equivalent normal-form~\cite{Kuhn53}, but since it is
exponentially larger, most algorithms have to be adapted
to handle the sequential setting directly.

There are several algorithms for computing strategies.
In zero-sum
games (where $\forall \pi \in \Pi, \vec{1} \cdot U(\pi) = 0)$,
one can use e.g. linear programming, fictitious play~\cite{Brown51},
replicator dynamics~\cite{TaylorJonkerRD}, or regret minimization~\cite{Blum07}.
Some of these techniques have been extended
to extensive (sequential) form~\cite{Heinrich15FSP,Gatti13Efficient,Lanctot14Further,CFR}
with an exponential increase in the size of the state space.
However, these extensions have almost exclusively treated the two-player case, with
some notable exceptions~\cite{Lanctot14Further,Gibson13}.
Fictitious play also converges in potential games which includes
cooperative (identical payoff) games.

\begin{figure}[t!]
\begin{center}
\includegraphics[scale=0.25]{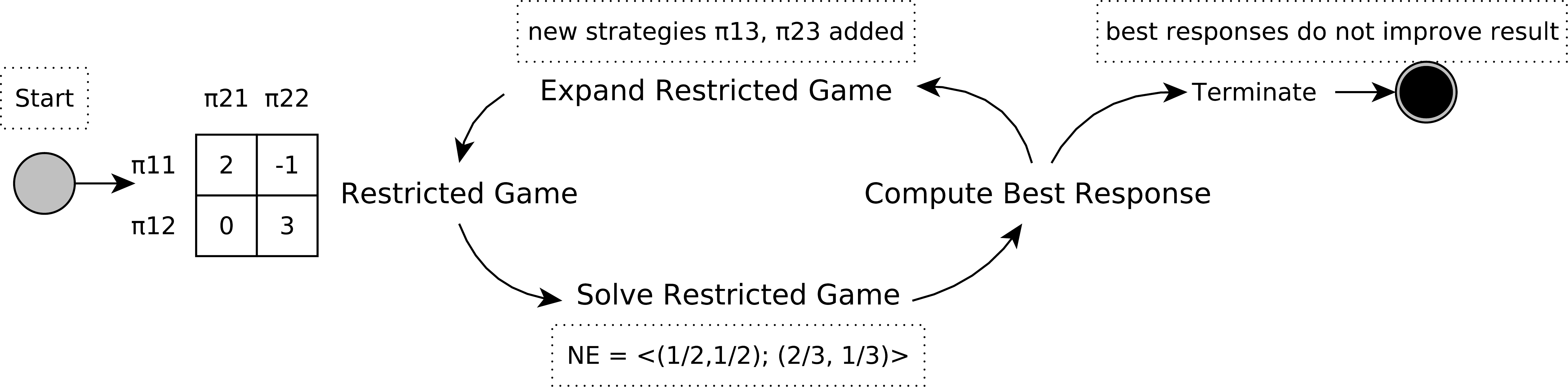}
\caption{The Double Oracle Algorithm. Figure taken from~\cite{Bosansky16AIJ} with authors' permission. \label{fig:do}}
\end{center}
\end{figure}

The {\bf double oracle} (DO) algorithm~\cite{McMahan03Planning}
solves a set of (two-player, normal-form) subgames induced by subsets $\Pi^t \subset \Pi$ at time $t$.
A payoff matrix for the subgame $G_t$ includes only those entries corresponding to the
strategies in $\Pi^t$. At each time step $t$, an equilibrium $\sigma^{*,t}$ is obtained for
$G^t$, and to obtain $G^{t+1}$ each player adds a best response $\pi^{t+1}_i \in \BR(\sigma^{*,t}_{-i})$ from the
full space $\Pi_i$, so for all $i$, $\Pi_i^{t+1} = \Pi_i^t \cup \{ \pi^{t+1}_i \}$.
The algorithm is illustrated in Figure~\ref{fig:do}.
Note that finding an equilibrium in a zero-sum game takes time polynomial in $|\Pi^t|$, and is
PPAD-complete for general-sum~\cite{Shoham09}.

Clearly, DO is guaranteed to converge to an equilibrium in two-player games.
But, in the worst-case, the entire strategy space may have to be enumerated.
For example, this is necessary for Rock-Paper-Scissors, whose only equilibrium has full
support $(\frac{1}{3}, \frac{1}{3}, \frac{1}{3})$. However, there is evidence that support
sizes shrink for many games as a function of episode length, how much hidden information is revealed
and/or effects it has on the payoff~\cite{Long10Understanding,SupportPaper,Bosansky16AIJ}.
Extensions to the extensive-form games have been
developed~\cite{Zinkevich07New,Bosansky13Using,Bosansky16AIJ} but still
large state spaces are problematic due to the curse of dimensionality.

Empirical game-theoretic analysis (EGTA) is the study of meta-strategies obtained through
simulation in complex games~\cite{Walsh02,Wellman06}. An \defword{empirical game}, much
smaller in size than the full game, is constructed by discovering strategies,
and meta-reasoning about the strategies to navigate the strategy space.
This is necessary when it is prohibitively expensive to explicitly
enumerate the game's strategies.
Expected utilities for each joint strategy are estimated and recorded in an
\defword{empirical payoff table}. The empirical game is analyzed, and the simulation process
continues. EGTA has been employed in trading agent competitions (TAC) and automated bidding
auctions.

One study used evolutionary dynamics in the space of known expert meta-strategies in
Poker~\cite{Ponsen09EGTPoker}.
Recently, reinforcement learning has been used to {\it validate} strategies found via
EGTA~\cite{Wright16}.
In this work, we aim to discover new strategies through learning. However, instead
of computing exact best responses, we compute approximate best responses using
reinforcement learning.
A few epochs of this was demonstrated in continuous double auctions
using tile coding~\cite{Schvartzman09}. This work follows up in this line,
running more epochs, using modern function approximators (deep networks),
a scalable implementation, and with a focus on finding policies that
can generalize across contexts.


A key development in recent years is deep learning \cite{LeCun15}.
While most work in deep learning has focused on
supervised learning, impressive results have recently been shown
using deep neural networks for reinforcement learning, e.g. \cite{Silver16AlphaGo,HeessWSLET15,
Mnih15DQN,DOhGLLS15}. For instance, Mnih et al. \cite{Mnih15DQN}
train policies for playing Atari video games and 3D navigation~\cite{mnih2016asynchronous},
given only screenshots.
Silver et al. introduced AlphaGo \cite{Silver16AlphaGo,Silver17AG0}, combining deep RL with Monte Carlo tree
search, outperforming human experts.


Computing approximate responses is more computationally feasible, and fictitious play
can handle approximations~\cite{Hofbauer02SFP,Leslie06GWFP}. It is also more
biologically plausible given natural constraints of bounded rationality. In
{\bf behavioral game theory}~\cite{Wright10,Wright17}, the focus is to {\it predict}
actions taken by humans, and the responses are intentionally constrained to
increase predictive ability. A recent work uses a deep learning architecture~\cite{Hartford16}.
The work that closely resembles ours is {\bf level-k} thinking~\cite{CGC06} where level $k$
agents respond to level $k-1$ agents, and more closely {\bf cognitive hierarchy}~\cite{Camerer04},
in which responses are to distributions over levels $\{ 0, 1, \ldots, k-1 \}$.
However, our goals and motivations are very different: we use the setup as a means
to produce more general policies rather than to predict human behavior. Furthermore, we consider
the sequential setting rather than normal-form games.

Lastly, there has been several studies from the literature on
co-evolutionary algorithms; specifically, how learning cycles
and overfitting to the current populations
can be mitigated~\cite{Oliehoek06Nash,ieeetevcssamot2012,Kouvaris17}.

\section{Policy-Space Response Oracles \label{sec:psro}}

We now present our main conceptual algorithm, policy-space response oracles (PSRO).
The algorithm is a natural generalization of Double Oracle where the meta-game's
choices are policies rather than actions. It also generalizes Fictitious
Self-Play~\cite{Heinrich15FSP,Heinrich16}. Unlike previous work, any meta-solver can
be plugged in to compute a new meta-strategy.
In practice, parameterized policies (function approximators)
are used to generalize across the state space without requiring any domain knowledge.

The process is summarized in Algorithm~\ref{alg:psro}.
The meta-game is represented as an empirical game, starting with a single
policy (uniform random) and growing, each epoch, by adding policies
(``oracles'') that approximate best responses to the meta-strategy of the other players.
In (episodic) partially observable multiagent environments, when the other players are fixed
the environment becomes Markovian and computing a best response reduces to solving a
form of MDP~\cite{Greenwald17}. Thus, any reinforcement learning algorithm can be used.
We use deep neural networks due to the recent success in reinforcement learning.
In each episode, one player is set to {\it oracle}(learning) mode to train
$\pi_i'$, and a fixed policy is sampled from the opponents' meta-strategies ($\pi_{-i} \sim \sigma_{-i}$).
At the end of the epoch, the new oracles are added to their policy sets $\Pi_i$,
expected utilities for new policy combinations are computed via simulation
and added to the empirical tensor $U^\Pi$, which takes time exponential in $|\Pi|$.

Define $\Pi^T = \Pi^{T-1} \cup \{ \pi' \}$ as the policy space including the currently learning oracles,
and $|\sigma_{i}| = |\Pi^T_{i}|$ for all $i \in [[n]]$.
Iterated best response is an instance of PSRO with $\sigma_{-i} = (0, 0, \cdots, 1, 0)$.
Similarly, Independent RL and fictitious play are instances of PSRO with $\sigma_{-i} = (0, 0, \cdots, 0, 1)$
and $\sigma_{-i} = (1/K, 1/K, \cdots, 1/K, 0)$, respectively, where $K = |\Pi_{-i}^{T-1}|$.
Double Oracle is an instance of PSRO with $n = 2$ and $\sigma^T$ set to a Nash equilibrium profile of
the meta-game $(\Pi^{T-1}, U^{\Pi^{T-1}})$.

\begin{figure}[t!]
\begin{minipage}[h]{6.7cm}
\vspace{0pt}
\begin{algorithm2e}[H]
\SetKwInOut{Input}{input}\SetKwInOut{Output}{output}
\Input{initial policy sets for all players $\Pi$}
Compute exp. utilities $U^\Pi$ for each joint $\pi \in \Pi$ \;
Initialize meta-strategies $\sigma_i = \textsc{Uniform}(\Pi_i)$ \;
\While{epoch $e$ in $\{ 1, 2, \cdots \}$}{
  \For{player $i \in [[n]]$}{
    \For{many episodes}{
      Sample $\pi_{-i} \sim \sigma_{-i}$ \;
      Train oracle $\pi_i'$ over $\rho \sim (\pi_i', \pi_{-i})$ \; 
    }
    $\Pi_i = \Pi_i \cup \{ \pi_i' \}$ \;
  }
  Compute missing entries in $U^\Pi$ from $\Pi$ \label{alg:psro:update_ept} \;
  Compute a meta-strategy $\sigma$ from $U^\Pi$ \label{alg:psro:metasol} \;
}
Output current solution strategy $\sigma_i$ for player $i$ \;
\caption{Policy-Space Response Oracles\label{alg:psro}}
\end{algorithm2e}
\end{minipage}
~~~~~~~
\begin{minipage}[h]{6.8cm}
\vspace{0pt}
\begin{algorithm2e}[H]
\SetKwInOut{Input}{input}\SetKwInOut{Output}{output}
\Input{player number $i$, level $k$}
\While{not terminated}{
    \textsc{CheckLoadMS}$(\{j | j \in [[n]], j \not= i \}, k)$ \;
    \textsc{CheckLoadOracles}$(j \in [[n]], k' \le k)$ \;
    \textsc{CheckSaveMS}$(\sigma_{i,k})$ \;
    \textsc{CheckSaveOracle}$(\pi_{i,k})$ \;
    Sample $\pi_{-i} \sim \sigma_{-i,k}$ \;
    Train oracle $\pi_{i,k}$ over $\rho_1 \sim (\pi_{i,k}, \pi_{-i})$ \;
    \If {iteration number $\mathbf{mod}~T_{ms} = 0$}{
      Sample $\pi_i \sim \sigma_{i,k}$ \;
      Compute $u_i(\rho_2)$, where $\rho_2 \sim (\pi_i, \pi_{-i})$ \;
      Update stats for $\pi_i$ and update $\sigma_{i,k}$
    }
}
Output $\sigma_{i,k}$ for player $i$ at level $k$
\caption{Deep Cognitive Hierarchies \label{alg:dch}}
\end{algorithm2e}
\end{minipage}
\end{figure}

An exciting question is what can happen with (non-fixed) meta-solvers outside this known space?
Fictitious play is agnostic to the policies it is responding to; hence it can only sharpen the meta-strategy
distribution by repeatedly generating the same best responses. On the other hand, responses to equilibrium
strategies computed by Double Oracle will (i) overfit to a specific equilibrium in the $n$-player or general-sum case,
and (ii) be unable to generalize to parts of the space not reached by any equilibrium strategy in the
zero-sum case. Both of these are undesirable when computing general policies that should work well in any context.
We try to balance these problems of overfitting with a compromise: meta-strategies with full support that force
(mix in) $\gamma$ exploration over policy selection.


\subsection{Meta-Strategy Solvers}

A meta-strategy solver takes as input the empirical game $(\Pi, U^\Pi)$ and produces a meta-strategy
$\sigma_i$ for each player $i$.
We try three different solvers: regret-matching, Hedge, and projected replicator dynamics.
These specific meta-solvers accumulate values for each policy (``arm'') and an aggregate value based on
all players' meta-strategies. We refer to $u_i(\sigma)$ as player $i$'s expected value given all players'
meta-strategies and the
current empirical payoff tensor $U^\Pi$ (computed via multiple tensor dot products.) Similarly, denote
$u_i(\pi_{i,k}, \sigma_{-i})$ as the expected utility if player $i$ plays their $k^{th} \in [[K]] \cup \{0\}$
policy and the other
players play with their meta-strategy $\sigma_{-i}$. Our strategies use an exploration parameter $\gamma$,
leading to a lower bound of $\frac{\gamma}{K+1}$ on the probability of selecting any $\pi_{i,k}$.

The first two meta-solvers (Regret Matching and Hedge) are straight-forward applications of previous
algorithms, so we defer the details to Appendix~\ref{sec:metasolvers}.
Here, we introduce a new solver we call \defword{projected replicator dynamics} (PRD).
From Appendix~\ref{sec:metasolvers}, when using the asymmetric replicator dynamics, \eg with two players,
where $U^\Pi = (\mathbf{A}, \mathbf{B})$, 
the change in probabilities for the $k^{th}$ component (\ie the policy $\pi_{i,k}$) of meta-strategies
$(\sigma_1, \sigma_2) = (\mathbf{x}, \mathbf{y})$ are:
\[
\frac{dx_k}{dt}= x_k[(\mathbf{A} \mathbf{y})_k-\mathbf{x}^T \mathbf{A} \mathbf{y}], ~~~~~~
\frac{dy_k}{dt}= y_k[(\mathbf{x}^T \mathbf{B})_k-\mathbf{x}^T \mathbf{B} \mathbf{y}],
\]
To simulate the replicator dynamics in practice, discretized updates are simulated using a step-size of $\delta$.
We add a projection operator $P(\cdot)$
to these equations that guarantees exploration:
$\mathbf{x} \leftarrow P( \mathbf{x} + \delta \frac{d \mathbf{x}}{dt}),$
$\mathbf{y} \leftarrow P( \mathbf{y} + \delta \frac{d \mathbf{y}}{dt}),$
where $P(\mathbf{x}) = \argmin_{\mathbf{x'} \in \Delta^{K+1}_\gamma}\{ ||\mathbf{x'} - \mathbf{x}|| \}$,
if any $x_k < \gamma/(K+1)$ or $\mathbf{x}$ otherwise, and
$\Delta^{K+1}_\gamma = \{ \mathbf{x}~|~x_k \ge \frac{\gamma}{K+1}, \sum_k{x_k} = 1 \}$ is the $\gamma$-exploratory simplex
of size $K+1$. This enforces exploratory $\sigma_i(\pi_{i,k}) \ge \gamma/(K+1)$. The PRD approach
can be understood as directing exploration in comparison to standard replicator dynamics approaches that contain isotropic diffusion or mutation terms (which assume undirected and unbiased evolution), for more details see \cite{Tuyls08}.

%
%
%
%

\subsection{Deep Cognitive Hierarchies}

%
\begin{wrapfigure}{rt}{0.4\textwidth}
  \centering
    \vspace{-0.75cm}
    \includegraphics[clip,width=0.37\textwidth]{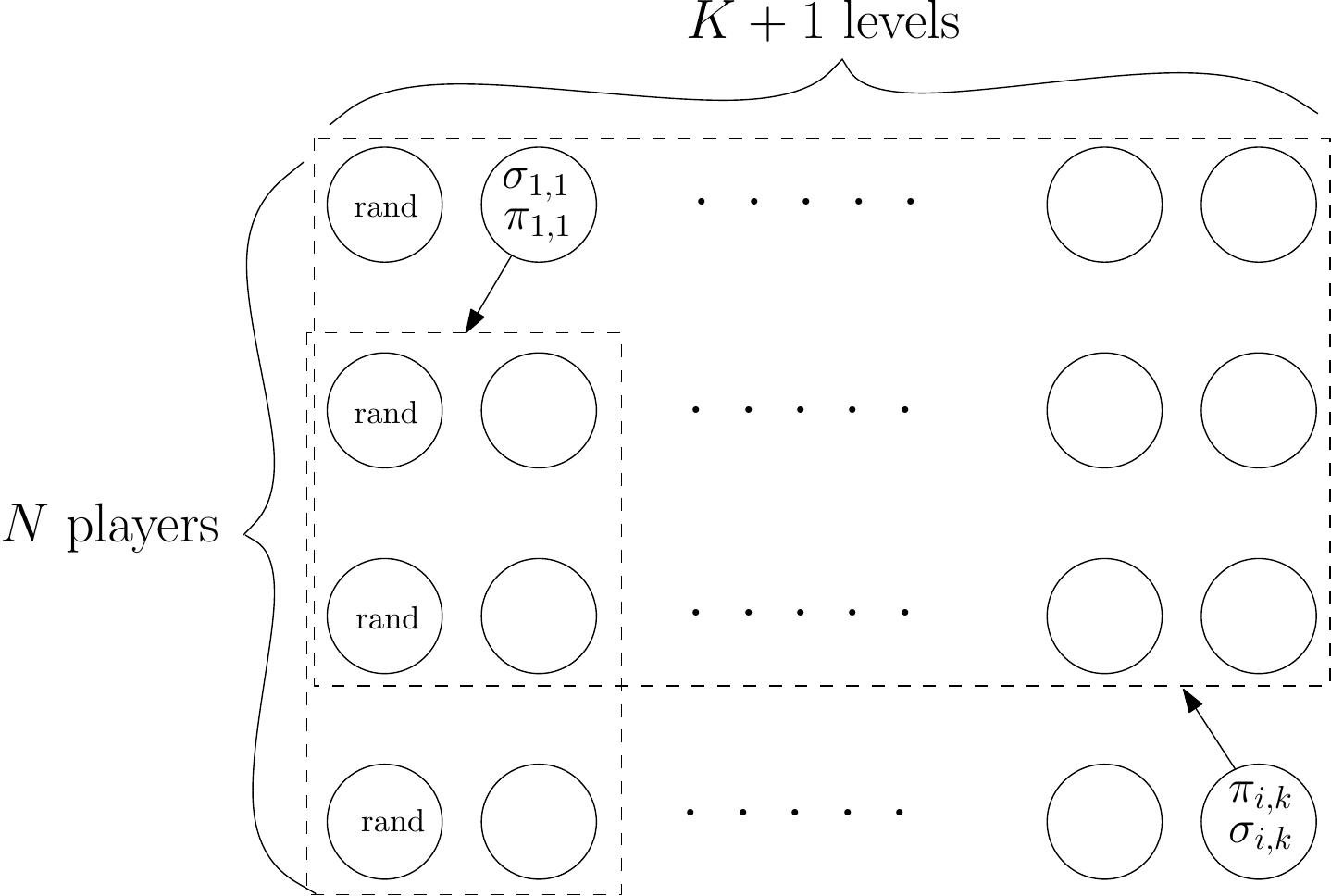}
  \caption{Overview of DCH \label{fig:dch}}
\end{wrapfigure}

While the generality of PSRO is clear and appealing, the RL step can take a long time to converge to a good response.
In complex environments, much of the basic behavior that was learned in one epoch may
need to be relearned when starting again from scratch; also, it may be desirable
to run many epochs to get oracle policies that can recursively reason through deeper levels of contingencies.

To overcome these problems, we introduce a practical parallel form of PSRO. Instead of an unbounded number of
epochs, we choose a fixed number of {\it levels} in advance.
Then, for an $n$-player game, we start $nK$ processes in parallel (level 0 agents are uniform random):
each one trains a single oracle policy $\pi_{i,k}$ for player $i$ and level $k$ and updates its own meta-strategy $\sigma_{i,k}$,
saving each to a central disk periodically.
Each process also maintains copies of all the other oracle policies $\pi_{j, k' \le k}$ at the current and lower levels,
as well as the meta-strategies at the current level $\sigma_{-i,k}$, which are periodically refreshed from a central disk.
We circumvent storing $U^\Pi$ explicitly by updating the meta-strategies online.
We call this a Deep Cognitive Hierarchy (DCH), in reference to Camerer, Ho, \& Chong's model augmented with deep RL.
Example oracle response dynamics are shown in Figure~\ref{fig:dch}, and pseudo-code in Algorithm~\ref{alg:dch}.

Since each process uses slightly out-dated copies of the other process's policies and meta-strategies, DCH
approximates PSRO. Specifically, it trades away accuracy of the correspondence to PSRO for practical efficiency and,
in particular, scalability.
Another benefit of DCH is
an asymptotic reduction in total space complexity. In PSRO, for $K$ policies and $n$ players, the space required to store
the empirical payoff tensor is $K^n$. Each process in DCH stores $nK$ policies of fixed size, and $n$ meta-strategies (and
other tables) of size bounded by $k \le K$. Therefore the total space required is $O(nK \cdot (nK + nK)) = O(n^2 K^2)$.
This is possible is due to the use of {\it decoupled} meta-solvers, which compute strategies online without
requiring a payoff tensor $U^\Pi$, which we describe now.

\subsubsection{Decoupled Meta-Strategy Solvers}

In the field of online learning, the
experts algorithms (``full information'' case) receive information about each arm at every round. In the
bandit (``partial information'') case, feedback is only given for the arm that was pulled.
Decoupled meta-solvers are essentially sample-based adversarial bandits~\cite{Bubeck12Bandits} applied to games.
Empirical strategies are known to converge to Nash equilibria in certain classes of games
(i.e. zero-sum, potential games) due to the folk theorem~\cite{Blum07}.

We try three: decoupled regret-matching~\cite{Hart01}, Exp3 (decoupled Hedge)~\cite{Exp3}, and decoupled PRD.
Here again, we use exploratory strategies with $\gamma$ of the uniform strategy mixed in, which is also necessary
to ensure that the estimates are unbiased.
For decoupled PRD, we maintain running averages for the overall average value an value of each arm (policy).
Unlike in PSRO, in the case of DCH, one sample is obtained at a time and the meta-strategy is updated periodically
from online estimates.

\section{Experiments}

In all of our experiments, oracles use Reactor~\cite{Gruslys17} for learning,
which has achieved state-of-the-art results in Atari game-playing. Reactor uses
Retrace$(\lambda)$~\cite{munos16safe}
for off-policy policy evaluation, and $\beta$-Leave-One-Out
policy gradient for policy updates, and supports recurrent network training, which
could be important in trying to match online experiences to those observed during training.

The action spaces for each player are identical, but the algorithms do not require this.
Our implementation differs slightly from the conceptual descriptions in Section~\ref{sec:psro};
see App.~\ref{sec:alg_details} for details.

{\bf First-Person Gridworld Games.} 
Each agent has a local field-of-view (making the world partially observable),
sees 17 spaces in front, 10 to either side, and 2 spaces behind. Consequently,
observations are encoded as 21x20x3 RGB tensors with values 0 -- 255.
Each agent has a choice of turning left or right, moving forward or backward, stepping left or
right, not moving, or casting an endless light beam in their current direction. In addition, the agent
has two composed actions of moving forward and turning. Actions are executed
simultaneously, and order of resolution is randomized.
Agents start on a random spawn point at the beginning of each episode.
If an agent is touched (``tagged'') by another agent's light beam twice, then the target agent is
immediately teleported to a spawn point. In {\it laser tag}, the source agent then receives 1 point of
reward for the tag.
In another variant, {\it gathering}, there is no tagging but agents can collect apples,
for 1 point per apple, which refresh at a fixed rate. In {\it pathfind}, there is no tagging
nor apples, and both agents get 1 point reward when both reach their destinations, ending the episode.
In every variant, an episode consists of 1000 steps of simulation. Other details,
such as specific maps, can be found in Appendix~\ref{sec:environments}.

{\bf Leduc Poker} is a common benchmark in Poker AI,
consisting of a six-card deck: two suits with three cards (Jack, Queen, King)
each. Each player antes 1 chip to play, and receives one private card.
There are two rounds of betting, with a maximum of two raises each, whose values
are 2 and 4 chips respectively. After the first round of betting, a single public
card is revealed.
The input is represented as in~\cite{Heinrich16}, which includes one-hot encodings of the private card,
public card, and history of actions.
Note that we use a more difficult version than in previous work;
see Appendix~\ref{sec:games_repr} for details.


\subsection{Joint Policy Correlation in Independent Reinforcement Learning \label{sec:jpc}}

To identify the effect of overfitting in independent reinforcement learners, we introduce
{\bf joint policy correlation} (JPC) matrices. To simplify the presentation, we describe here
the special case of symmetric two-player games with non-negative rewards; for a general description,
see Appendix~\ref{sec:np_jpc}.

Values are obtained by running $D$ instances of the same experiment, differing only in the
seed used to initialize the random number generators.
Each experiment $d \in [[D]]$ (after many training episodes) produces policies $(\pi_1^d, \pi_2^d)$.
The entries of each $D \times D$ matrix shows the mean return over $T = 100$ episodes,
$\sum_{t=1}^T\frac{1}{T}(R_1^t + R_2^t)$, obtained when player 1 uses row policy $\pi_1^{d_i}$ and
and player 2 uses column policy $\pi_2^{d_j}$.
Hence, entries on the diagonals represent returns for policies that learned together (\ie same instance),
while off-diagonals show returns from policies that trained in separate instances.

\begin{figure}[h]
\centering
\begin{tabular}{ccc}
\includegraphics[width=0.4\textwidth]{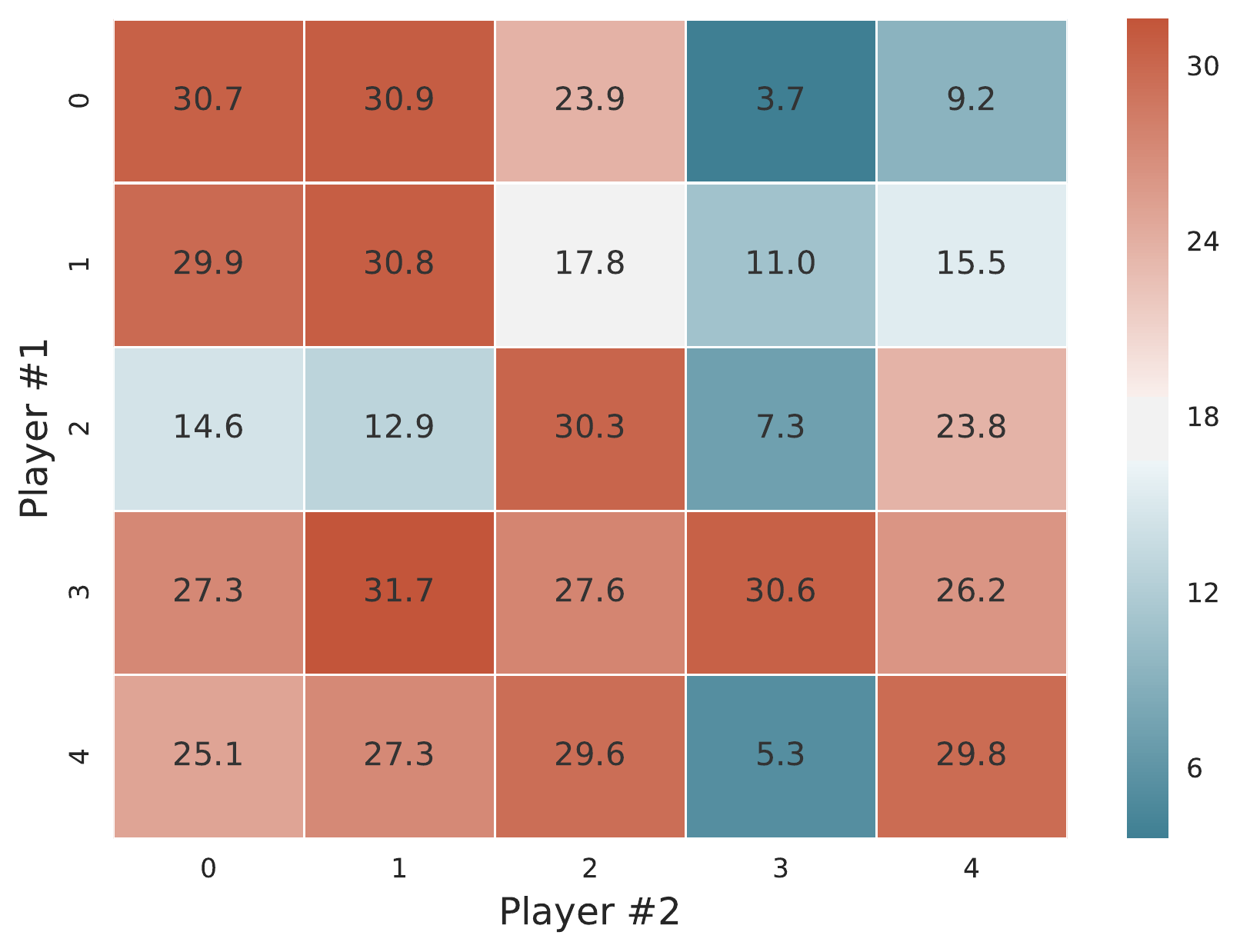} &
~~~ & \includegraphics[width=0.4\textwidth]{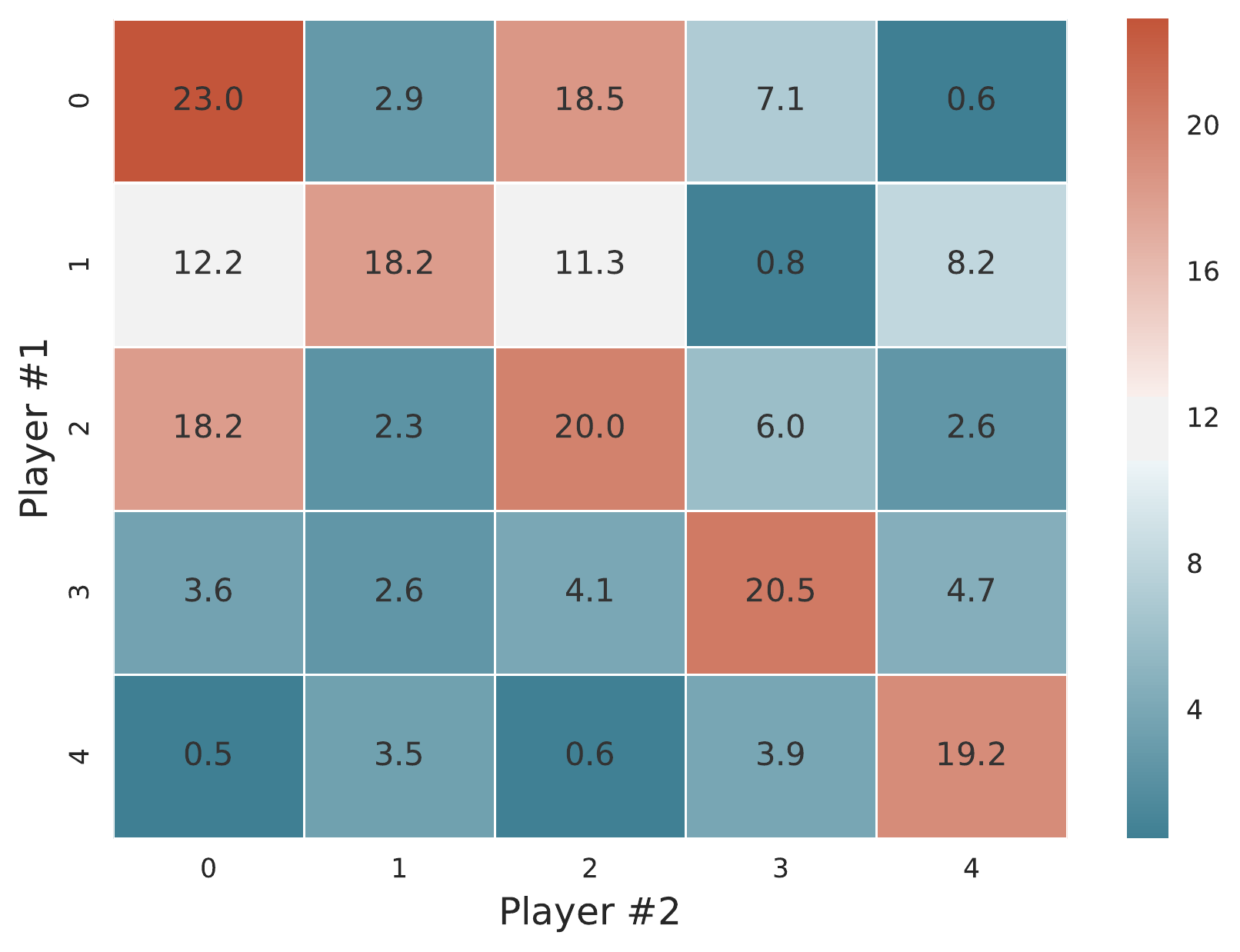} \\
\end{tabular}
\caption{Example JPC matrices for InRL on Laser Tag small2 map (left) and small4 (right). \label{fig:jpc_indqn_laser_tag}}
\end{figure}

From a JPC matrix, we compute an {\bf average proportional loss} in reward
as $R_- = (\bar{D}-\bar{O})/\bar{D}$ where $\bar{D}$
is the mean value of the diagonals and $\bar{O}$ is the mean value of the off-diagonals. E.g. in
Figure~\ref{fig:jpc_indqn_laser_tag}: $D = 30.44, O = 20.03, R_- = 0.342$.
Even in a simple domain with almost full observability (small2),
an independently-learned policy could expect to lose $34.2$\% of its reward when
playing with another
independently-learned policy even though it {\it was trained under identical circumstances}!
This clearly demonstrates an important problem with independent learners.
In the other variants (gathering and pathfind), we observe no JPC problem, presumably because
coordination is not required and the policies are independent. Results are summarized in Table~\ref{tbl:jpc}.
We have also noticed similar effects when using DQN~\cite{Mnih15DQN} as the oracle training algorithm;
see Appendix~\ref{sec:jpc_vids} for example videos.


\begin{table}[t]
\begin{center}
\begin{tabular}{|l|l|ccc|ccc|c|}
\hline
Environment       & Map       & \multicolumn{3}{|c|}{InRL} & \multicolumn{3}{|c|}{DCH(Reactor, 2, 10)} & JPC Reduction \\
                  &           &  $\bar{D}$  & $\bar{O}$  & $R_-$  & $\bar{D}$   & $\bar{O}$   & $R_-$ &       \\
\hline
Laser Tag         & small2    & 30.44 & 20.03 & 0.342  & 28.20   & 26.63  & 0.055 & 28.7 \% \\
Laser Tag         & small3    & 23.06 & 9.06  & 0.625  & 27.00   & 23.45  & 0.082 & 54.3 \% \\
Laser Tag         & small4    & 20.15 & 5.71  & 0.717  & 18.72   & 15.90  & 0.150 & 56.7 \% \\
Gathering         & field     & 147.34 & 146.89 & 0.003 & 139.70 & 138.74 & 0.007 &  --     \\
Pathfind          & merge     & 108.73 & 106.32 & 0.022 & 90.15  & 91.492 &  < 0  &  --     \\
\hline
\end{tabular}
\caption{Summary of JPC results in first-person gridworld games.\label{tbl:jpc}}
\end{center}
\end{table}

We see that a (level 10) DCH agent reduces the JPC problem significantly. On small2, DCH reduces the expected loss
down to 5.5\%, 28.7\% lower than independent learners.
The problem gets larger as the map size grows and problem becomes more partially observed,
up to a severe 71.7\% average loss. The reduction achieved by DCH also grows from 28.7\% to 56.7\%.

{\bf Is the Meta-Strategy Necessary During Execution?}
The figures above represent the fully-mixed strategy $\sigma_{i,10}$. We also analyze JPC
for only the highest-level policy $\pi_{i,10}$ in the laser tag levels.
The values are larger here: $R_- = 0.147, 0.27, 0.118$ for small2-4 respectively, showing the
importance of the meta-strategy. However, these are still significant reductions in JPC: 19.5\%, 36.5\%, 59.9\%.

{\bf How Many Levels?}
On small4, we also compute values for level 5 and level 3: $R_- = 0.156$ and $R_- = 0.246$,
corresponding to reductions in JPC of 56.1\% and 44\%, respectively. Level 5 reduces JPC by a similar amount
as level 10 (56.1\% vs 56.7\%), while level 3 less so (44\% vs. 56.1\%.)




\subsection{Learning to Safely Exploit and Indirectly Model Opponents in Leduc Poker}


We now show results for a Leduc poker where strong benchmark algorithms exist, such as
counterfactual regret (CFR) minimization~\cite{CFR,Bowling15Poker}.
We evaluate our policies using two metrics: the first is performance against fixed players
(random, CFR's average strategy after 500 iterations ``cfr500'',
and a purified version of ``cfr500pure'' that chooses the action with highest probability.)
The second is commonly used in poker AI:
\textsc{NashConv}$(\sigma) = \sum_i^n \max_{\sigma_i' \in \Sigma_i} u_i(\sigma_i', \sigma_{-i}) - u_i(\sigma)$,
representing, in total, how much each player gains by deviating to their best response (unilaterally),
a value that can be interpreted as a distance from a Nash equilibrium (called {\bf exploitability} in the two-player setting).
NashConv is easy to compute in small enough games~\cite{2011-ijcai-abr}; for CFR's values see Appendix~\ref{sec:expl_cfr}.

\begin{figure}[b!]
\begin{tabular}{cc}
    \includegraphics[width=0.62\textwidth]{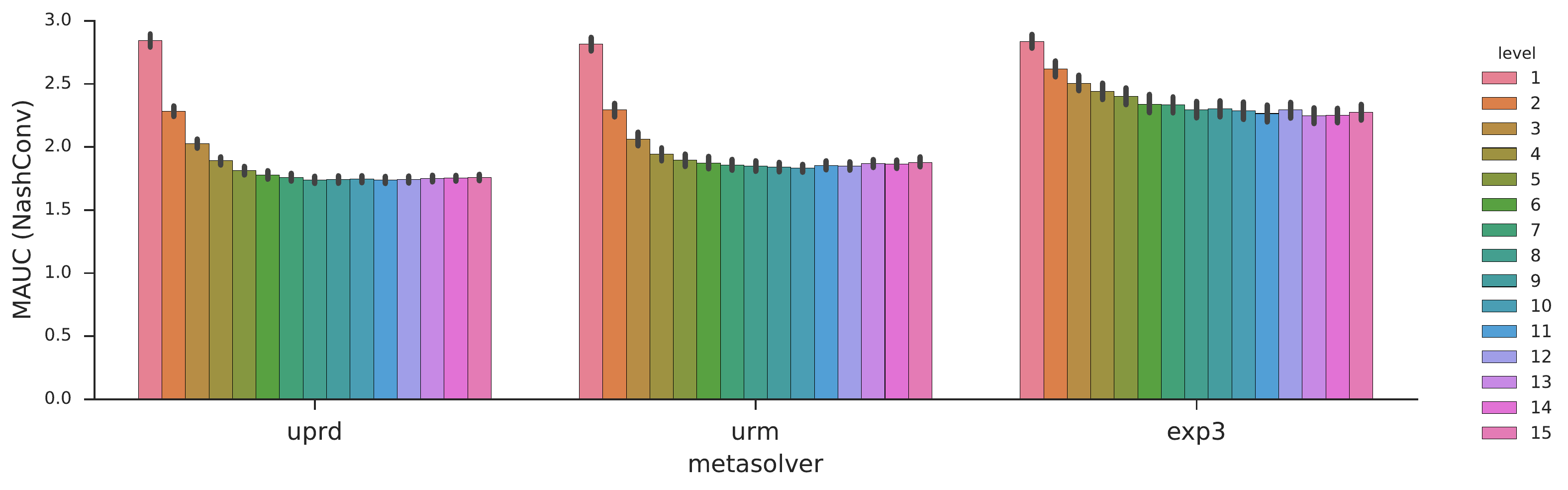} &
    \includegraphics[width=0.33\textwidth]{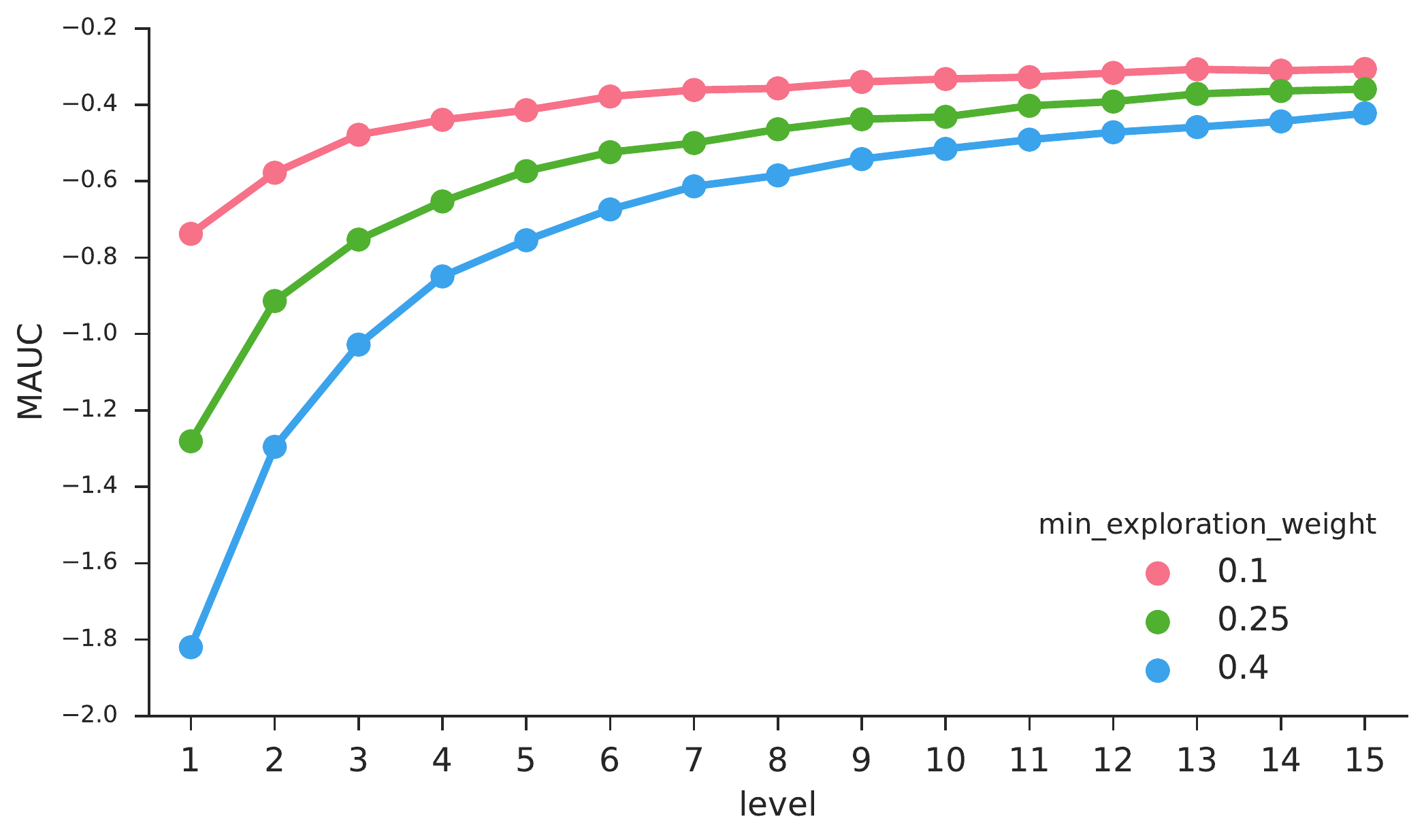}\\
    (a) & (b) \\
\end{tabular}
    \caption{(a) Effect of DCH parameters on NashConv in 2 player Leduc Poker. Left: decoupled PRD, Middle: decoupled RM, Right: Exp3, and
    (b) MAUC of the exploitation graph against cfr500.}
    \label{fig:dhc_params_2pleduc_1line}
\end{figure}

{\bf Effect of Exploration and Meta-Strategy Overview}.
We now analyze the effect of the various meta-strategies and exploration parameters.
In Figure~\ref{fig:dhc_params_2pleduc_1line}, we measure the mean area-under-the-curve (MAUC) of the NashConv
values for the last (right-most) 32 values in the NashConv graph, and exploration rate of $\gamma = 0.4$.
Figures for the other values of $\gamma$ are in Appendix~\ref{sec:app_results}, but we found this value
of $\gamma$ works best for minimizing NashConv. Also, we found that decoupled replicator dynamics
works best, followed by decoupled regret-matching and Exp3. Also, it seems that the higher the
level, the lower the resulting NashConv value is, with diminishing improvements.
For exploitation, we found that $\gamma=0.1$ was best, but the meta-solvers seemed to have little effect (see
Figure~\ref{fig:dhc_params_2pleduc_exploit}.)

\begin{figure}[h!]
  \centering
  \begin{subfigure}[b]{.49\textwidth}
    \includegraphics[width=\textwidth]{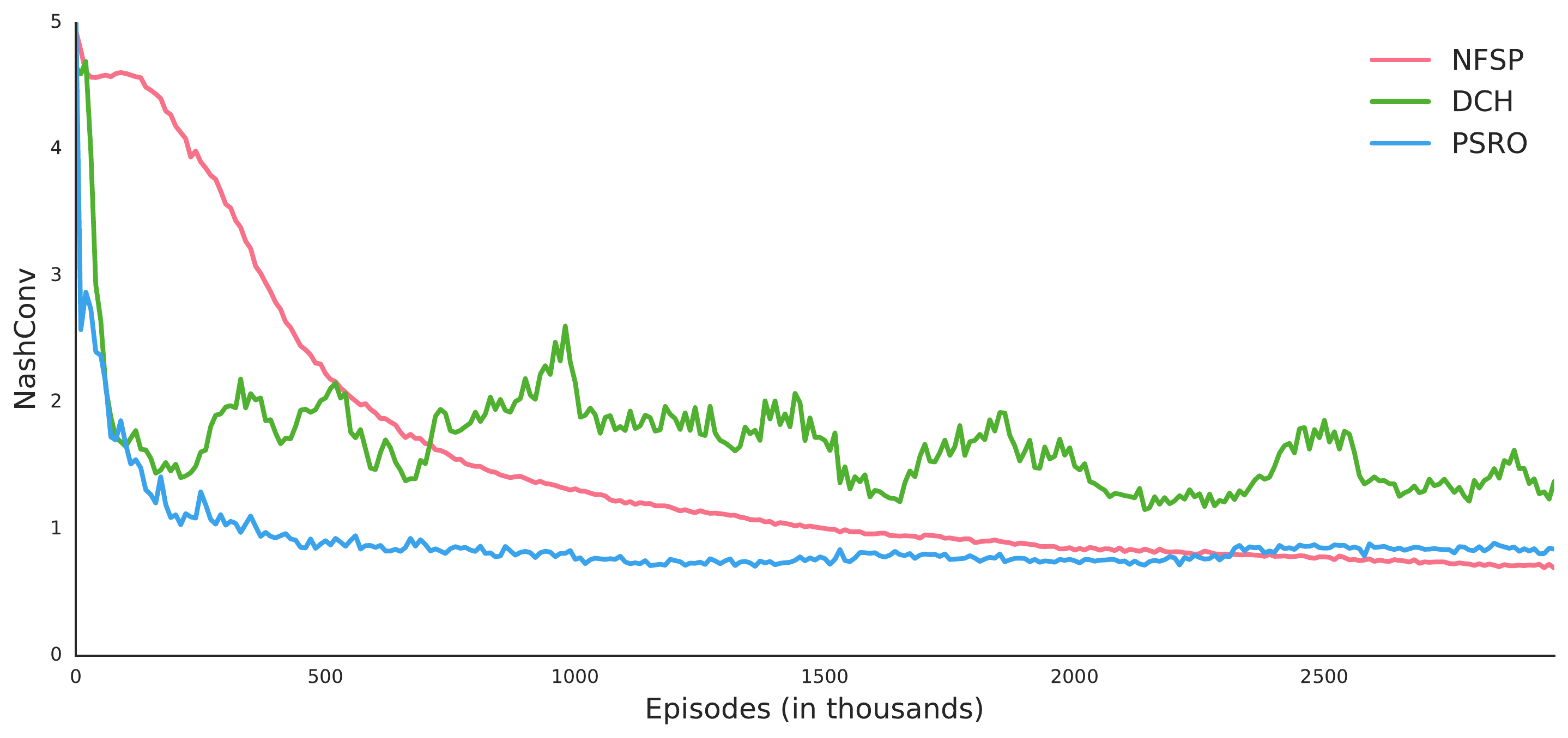}
    \caption{2 players}
    \label{fig:expl_2pleduc}
  \end{subfigure}
  \begin{subfigure}[b]{.49\textwidth}
    \includegraphics[width=\textwidth]{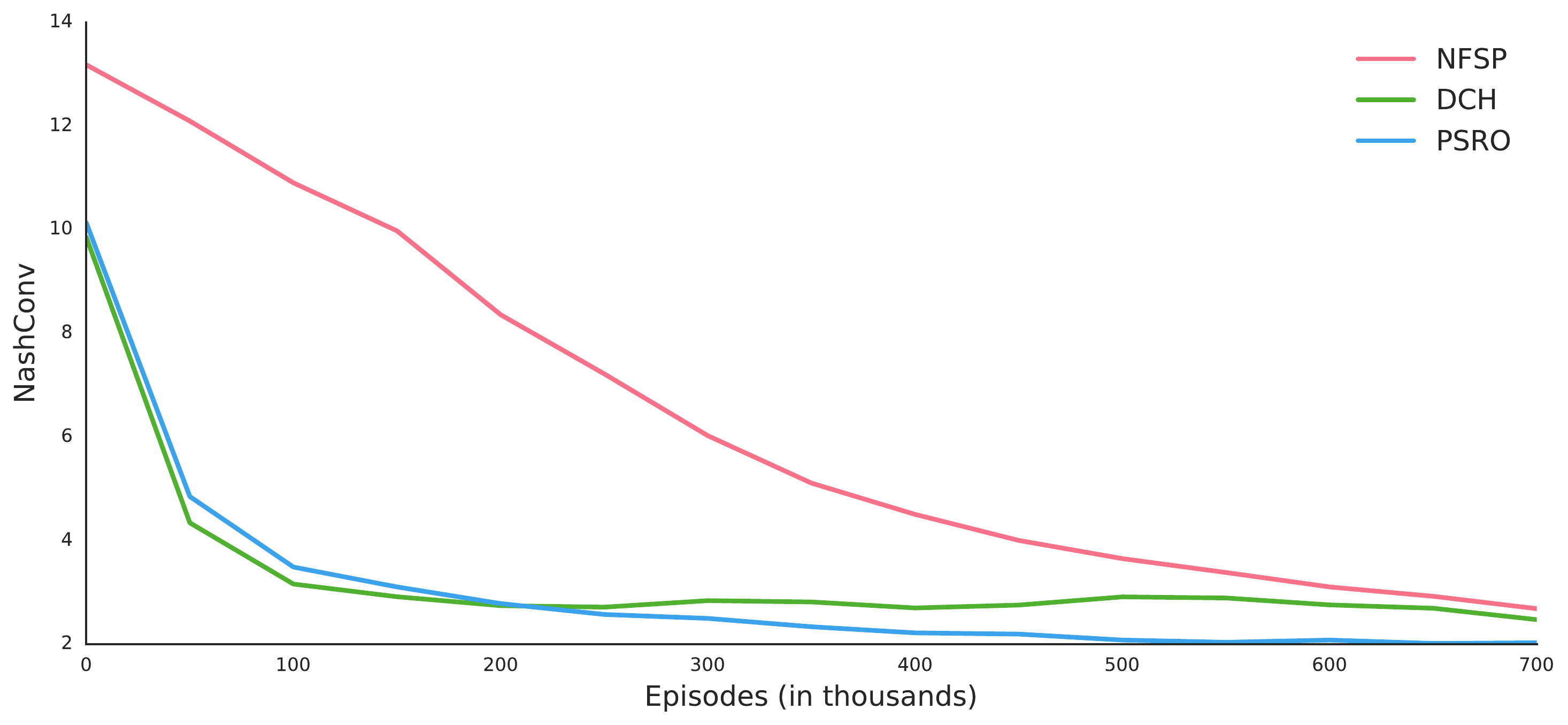}
    \caption{3 players}
    \label{fig:expl_3pleduc}
  \end{subfigure}

  \caption{Exploitability for NFSP x DCH x PSRO.}
  \label{fig:expl_leduc}
\end{figure}

\begin{figure}[h!]
  \centering
  \begin{subfigure}[b]{.32\textwidth}
    \includegraphics[width=\textwidth]{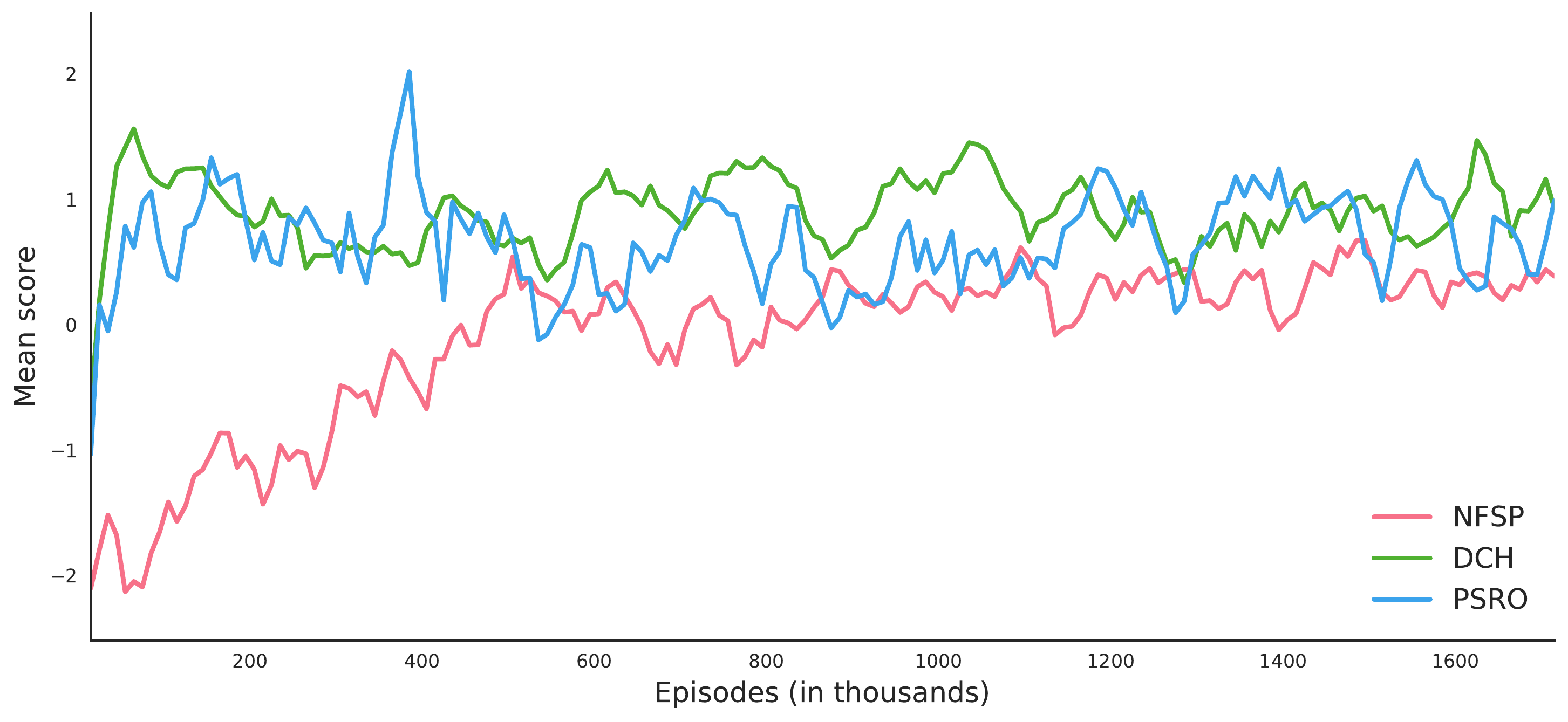}
    \caption{Random bots as ref. set}
    \label{fig:bots_rdn_2pleduc}
  \end{subfigure}
  \begin{subfigure}[b]{.32\textwidth}
    \includegraphics[width=\textwidth]{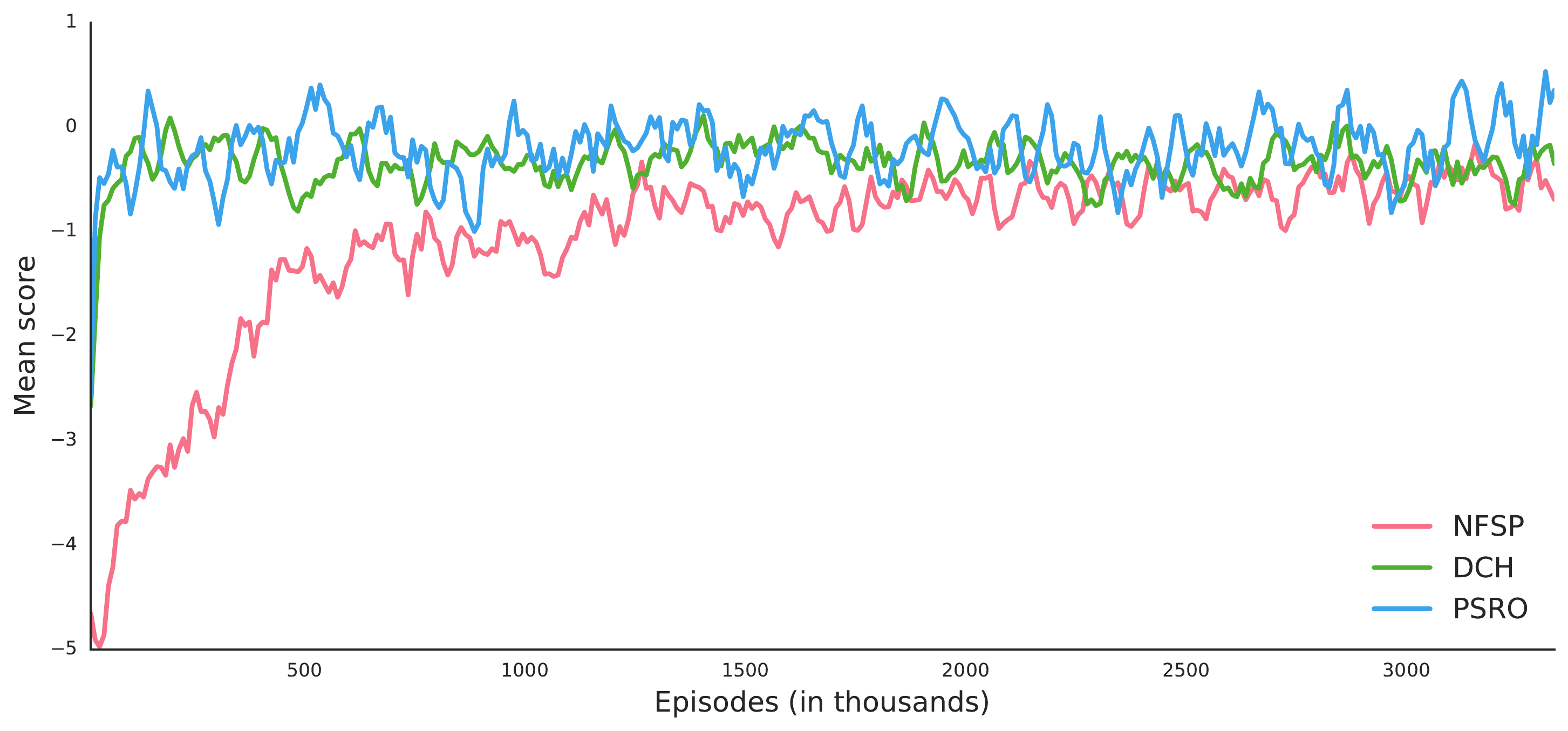}
    \caption{2-player CFR500 bots as ref. set}
    \label{fig:bots_cfr50_2pleduc}
  \end{subfigure}
  \begin{subfigure}[b]{.32\textwidth}
    \includegraphics[width=\textwidth]{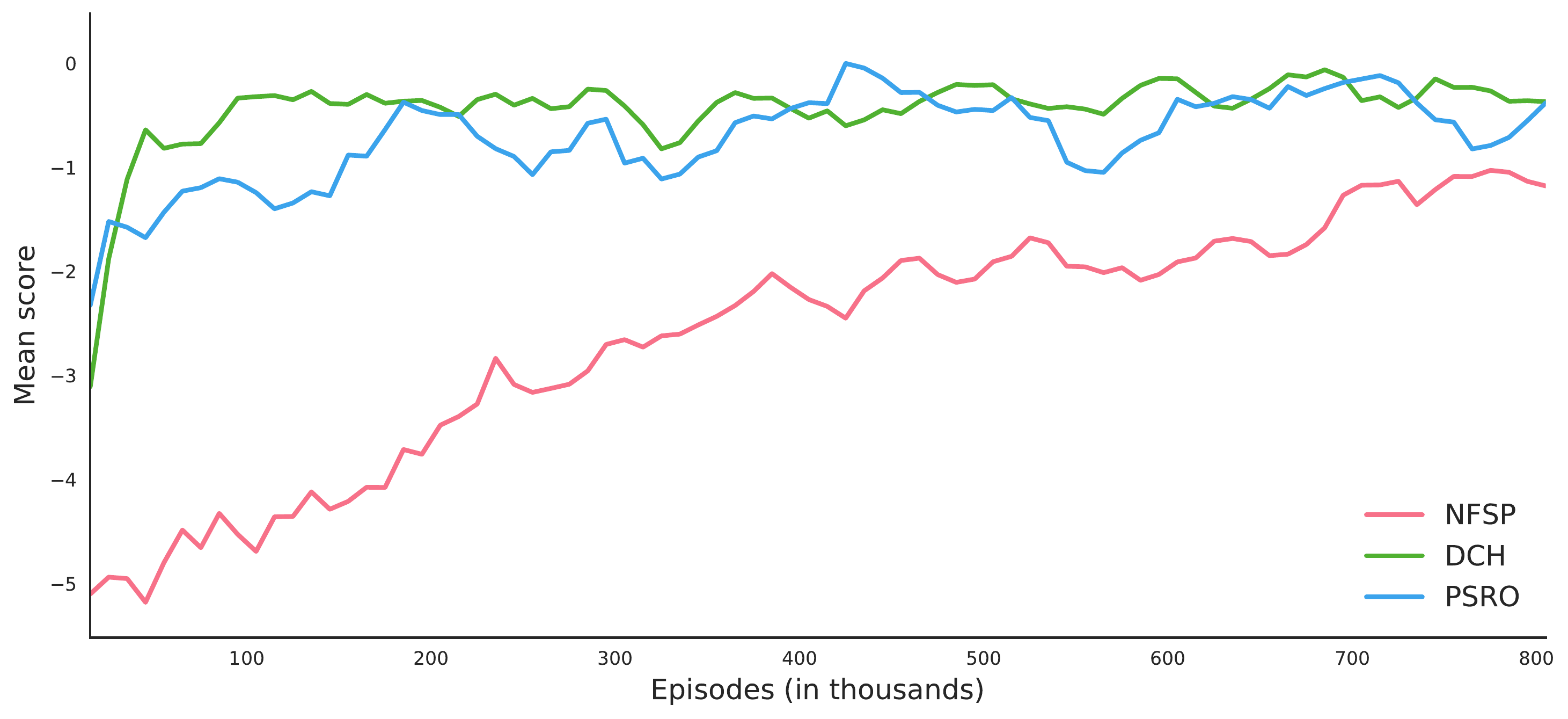}
    \caption{3-player CFR500 bots as ref. set}
    \label{fig:bots_cfr50_2pleduc}
  \end{subfigure}

  \caption{Evaluation against fixed set of bots. Each data point is an average of the four latest values.}
  \label{fig:bots_leduc2p}
\end{figure}

{\bf Comparison to Neural Fictitious Self-Play}.
We now compare to Neural Fictitious
Self-Play (NFSP)~\cite{Heinrich16}, an implementation of fictitious play
in sequential games using reinforcement learning.
Note that NFSP, PSRO, and DCH are all sample-based
learning algorithms that use general function approximation, whereas CFR is a tabular
method that requires a full game-tree pass per iteration.
NashConv graphs are shown for \{2,3\}-player in Figure~\ref{fig:expl_leduc}, and
performance vs. fixed bots in Figure~\ref{fig:bots_leduc2p}.

We observe that DCH (and PSRO) converge faster than NFSP at the start
of training, possibly due to a better meta-strategy than the uniform random one used in fictitious play.
The convergence curves eventually plateau: DCH in two-player is most affected, possibly due to the asynchronous
nature of the updates, and NFSP converges to a lower exploitability in later episodes.
We believe that this is due to NFSP's ability to learn a more accurate mixed average strategy at
states far down in the tree, which is particularly important in poker, whereas DCH and PSRO mix at the top
over full policies.

On the other hand, we see that PSRO/DCH are able to achieve higher performance against the fixed
players.
Presumably, this is because the policies produced by PSRO/DCH are better able to 
recognize flaws in the weaker opponent's policies, since the oracles are specifically trained
for this, and dynamically adapt to the exploitative response during the episode. So, 
NFSP is computing a safe equilibrium while
PSRO/DCH may be trading convergence precision for the ability to adapt to a range of different
play observed during training, in this context computing a robust counter-strategy~\cite{07nips-rnash,Ganzfried15}.







\section{Conclusion and Future Work}

In this paper, we quantify a severe problem with independent reinforcement learners,
joint policy correlation (JPC), that limits the generality of these approaches.
We describe a generalized algorithm for multiagent reinforcement learning 
that subsumes several previous algorithms.
In our experiments, we show that PSRO/DCH produces general policies that
significantly reduce JPC in partially-observable coordination games, and robust
counter-strategies that safely exploit opponents in a common competitive imperfect information game.
The generality offered by PSRO/DCH can be seen as a form of ``opponent/teammate regularization'', and has
also been observed recently in practice~\cite{Lowe17,Bansal17}.
We emphasize the game-theoretic foundations of these techniques, which we hope
will inspire further investigation into algorithm development for multiagent reinforcement learning.

In future work, we will consider maintaining diversity among oracles via loss penalties based on policy
dissimilarity, general response graph topologies, environments such as emergent language
games~\cite{Lazaridou:etal:2017} and RTS games~\cite{Tavares16,Sailer07}, and 
other architectures for prediction of behavior, such as opponent modeling~\cite{He16OMDRL}
and imagining future states via auxiliary tasks~\cite{JMC16UNREAL}.
We would also like to investigate fast online adaptation~\cite{AlShedivat17,Finn17} and
the relationship to computational Theory of Mind~\cite{Yoshida08,Baker11}, as well as
generalized (transferable) oracles over similar opponent policies using successor
features~\cite{Barreto17}.

{\bf Acknowledgments.} We would like to thank DeepMind and Google for providing an excellent research environment
that made this work possible. Also, we would like to thank the anonymous reviewers and several people for helpful comments:
Johannes Heinrich, Guy Lever, Remi Munos, Joel Z. Leibo, Janusz Marecki, Tom Schaul, Noam Brown, Kevin Waugh, Georg Ostrovski,
Sriram Srinivasan, Neil Rabinowitz, and Vicky Holgate.


\bibliographystyle{plain}
{\small \bibliography{psro}}

\newpage

\appendix
{\Large {\bf Appendices}}


\section{Meta-Solvers \label{sec:metasolvers}}

\subsection{Regret matching}

Regret matching (RM) is a simple adaptive procedure that leads to correlated
equilibria~\cite{Hart00}, which tabulates cumulative regret $R_i(\pi_{i,k})$ for $\pi_{i,k}$ at epoch
$k$. At each step, for all $i$ simultaneously: $R_i(\pi_{i,k}) \leftarrow R_i(\pi_{i,k}) + u_i(\pi_{i,k}, \sigma_{-i}) - u_i(\sigma)$.
A new meta-strategy is obtained by normalizing the positive portions $R_i$, and setting the
negative values to zero:
\[
\sigma_i(\pi_{i,k}) = \frac{R^+(\pi_{i,k})}{\sum_k^{[[K]] \cup \{0\}}  R^+(\pi_{i,k})}
~~\mbox{ if the denominator is positive},~~~ \mbox{ or } ~ \frac{1}{K+1} ~ \mbox{ otherwise },
\]
where $x^+ = \max(0,x)$.
In our case, we use exploratory strategies that enforce exploration: $\sigma_i' = \gamma \textsc{Unif}(K+1) + (1-\gamma) \sigma_i$.

\subsection{Hedge}

Hedge is similar, except it accumulates only rewards in $x_{i,k}$ and uses a softmax function to derive a
new strategy~\cite{Exp3}. At each step, $x_{i,k} \leftarrow x_{i,k} + u_i(\pi_{i,k}, \sigma_-i)$, and a new strategy
$\sigma_i(\pi_{i,k}) = \exp( \frac{\gamma}{K+1} x_{i,k} ) / \sum_k^{[[K]] \cup \{0\}} \exp(\frac{\gamma}{K+1} x_{i,k} )$.
Again here, we use strategies that mix in $\gamma \textsc{Unif}(K+1)$.

\subsection{Replicator Dynamics}

Replicator dynamics are a system of differential equations that describe how a population
of strategies, or replicators, evolve through time. In their most basic form they correspond to the biological \emph{selection}
principle, comparing the fitness of a strategy to the average fitness of the entire population. More specifically the symmetric
replicator dynamic mechanism is expressed as
\[
\frac{dx_k}{dt}=x_k[(\mathbf{A}\mathbf{x})_k-\mathbf{x}^T \mathbf{A} \mathbf{x}].
\]
Here, $x_k$ represents the density of strategy $\pi_{i,k}$ in the population ($\sigma_i(\pi_{i,k})$ for a given $k$), $A$ is the
payoff matrix which describes the different
payoff values each individual replicator receives when interacting with other replicators in the population. This common formulation
represents symmetric games, and typical examples of dynamics are taken from prisoner's dilemma, matching pennies, and stag hunt games.
For a more elaborate introduction to replicator dynamics, and their relationship with RL, we refer to \cite{BloembergenTHK15}.

Asymmetric replicator dynamics are applied to $n$-player games normal form games, \eg in two players using payoff tables $A$ and $B$, where $A\not=B^T$.
Examples are the infamous prisoner's dilemma and the Rock-Scissors-Paper games, in which both agents are interchangeable.
In the evolutionary setting this means that the agents are drawn from a single population. In general however, the symmetry assumption no longer
holds, as players do not necessarily have access to the same sets of strategies.
In this context it means we now have two players that come from different populations: 
\begin{equation*}\label{eq:P1rd}
 \frac{dx_k}{dt}= x_i[(A\mathbf{y})_k-\mathbf{x}^T A \mathbf{y}], ~~~~~~
 \frac{dy_k}{dt}= y_i[(\mathbf{x}^T B)_k-\mathbf{x}^T B\mathbf{y}],
\end{equation*}
where $x$ corresponds to the row player and $y$ to the column player.
In general, there are $n$ tensors, representing the utility to each player for each outcome.

In this paper we use a new {\it projected replicator dynamics} that enforces exploration by putting
a lower bound on the probability on $x_k$ and $y_k$.

\section{Joint Policy Correlation \label{sec:app_jpc}}

\subsection{Example Comparison Videos \label{sec:jpc_vids}}

This section points to several example videos of coordination (diagonals of the JPC experiments),
and miscoordination (off-diagonals of the JPC experiments).

{\bf Laser Tag (small2)}

  \begin{itemize}
  \item Diagonal: \url{https://www.youtube.com/watch?v=8vXpdHuoQH8}
  \item Off-Diagonal: \url{https://www.youtube.com/watch?v=jOjwOkCM_i8}
  \end{itemize}

{\bf Laser Tag (small3)}

  \begin{itemize}
  \item Diagonal: \url{https://www.youtube.com/watch?v=Z5cpIG3GsLw}
  \item Off-Diagonal: \url{https://www.youtube.com/watch?v=zilU0hXvGK4}
  \end{itemize}

In these videos, DQN was used to train the oracle policies, though the effects are similar
with ReActor.

\subsection{JPC in General $n$-player Environments \label{sec:np_jpc}}

In Section~\ref{sec:jpc} we introduced joint policy correlation for symmetric games with $n=2$
and non-negative rewards. In this section, we present the general description of JPC that
can be used for any finite (symmetric or asymmetric) $n$-player game with arbitrary rewards.

In general, each player has their own tensor of utilities values $U_i$. If there are $d$ separate
independent learning instances, then $U_i$ has dimensionality $d^n$ for each player $i$, and each
entry corresponds to an expected utility to player $i$ receives with a given combination of policies
produced for each player in each instance.

For example, for a four-player game and five independent learning instances (labeled $0, 1, \cdots 4$),
the value $U_4[0][3][2][2]$ corresponds to the expected return of the fourth player when:
\begin{itemize}
\item the first player uses their learned policy $\pi_1$ from instance 0,
\item the second player uses their learned policy $\pi_2$ from instance 3,
\item the third player uses their learned policy $\pi_3$ from instance 2,
\item the fourth player uses their learned policy $\pi_4$ from instance 2.
\end{itemize}

The definitions of average values over diagonals and off-diagonals and average proportional reduction
must now be indexed by the player $i$, and operate only on player $i$'s values in $U_i$. As a result, 
$\bar{D}_i$ is an average over $d$ values and $\bar{O}_i$ is an average over $d^n - d$ values, and $\bar{R}_i$
is defined analogously as in Section~\ref{sec:jpc} but instead using $\bar{D}_i$ and $\bar{O}_i$.
When $n$ is large an estimate of $\bar{O}_i$ can be used instead, by sampling $O(d)$ entries from the
exponentially many values.

Note that in asymmetric games, the JPC problem will vary across players. Therefore, it is not clear how to
aggregate and report a single value (summary). The simplest solution is to present a vector, $\vec{R}$, containing
$\bar{R}_i$ for each player, which exposes how each player is affected separately.

\section{Algorithm Details and Parameter Values \label{sec:alg_details}}

\subsection{Network, Oracle, and Training Parameters \label{sec:alg_training_details}}

Unless otherwise stated, we use the default parameter settings and architecture reported
in the Reactor paper. 
We set $\lambda = 0.9$, use a LSTM sequence (unroll) length of 32, with a batch size of 4,
learning rate $\alpha = 0.0001$, and momentum $\beta_1 = 0$, the ADAM optimizer~\cite{kingma2014adam},
replay buffer of size $500000$, and memorizing behavior probabilities for Retrace off-policy corrections.

\begin{figure}[h!]
\begin{center}
\includegraphics[width=0.4\textwidth]{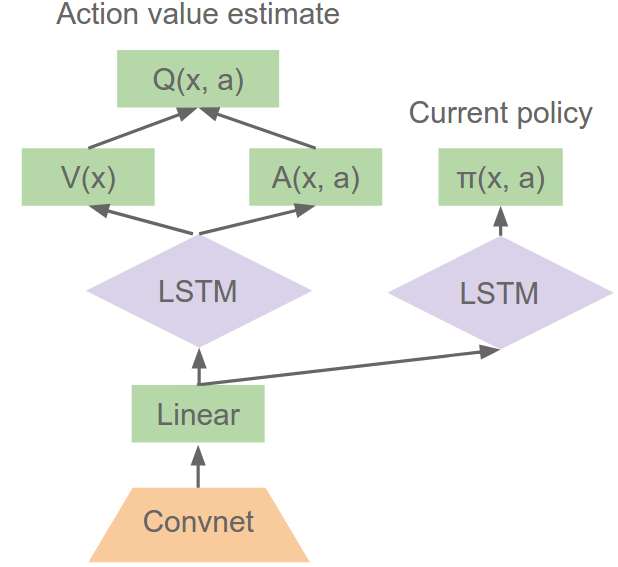}
\caption{Reactor network architecture. Modified file taken from~\cite{Gruslys17} with permission
of authors. \label{fig:reactor_arch}}
\end{center}
\end{figure}

The main network architecture is based on the default Reactor network,
except without a head for estimated behavior distributions (using purely memorized behavior probabilities),
as illustrated in Figure~\ref{fig:reactor_arch}.

For gridworld coordination games, we use three convolutional layers of
kernel widths (4, 5, 3) and strides (2, 1, 1) each outputting 8 planes.
The main fully-connected layer
has 32 units and each LSTM layer has 32 units.
Every layer, except the LSTM, was followed by a concatenated ReLU layer~\cite{ConcatRelu}
effectively doubling the number of outputs to 16 for following layers.
The rest of the architecture is the same as the default
architecture in the Reactor paper.

For Leduc poker, we use two fully-connected hidden layers of size 128, with rectified
linear units non-linearities at the end of each layer.
In each case, these are followed by an LSTM layer of size 32 and fully-connected layer of size 32.

\subsection{Neural Fictitious Self-Play (NFSP)}

In our tuning experiments for NFSP, we tried and found the following best values for NFSP.

Values of the form $x \rightarrow y$ refer to a linear schedule starting at $x$, ending at (and remaining at) $y$.
Values expresses as 5e-5 refer to $5 \cdot 10^{-5}$.

\subsubsection{NFSP parameters for in Two-Player Leduc}

\begin{center}
\begin{tabular}{|l|c|c|c|}
\hline
Parameter     & Value Set & Final Value \\
\hline
RL learning rate              & 1e-3, 1e-4, 1e-5 & 1e-3   \\
SL learning rate              & 1e-3, 1e-4, 1e-5 & 1e-5   \\
Anticipatory parameter        & 1e-3, 1e-4, 1e-5 & 0.1    \\
Reservoir buffer size         & 100000, 1000000 & 100000 \\
\hline
\end{tabular}\\
Best values for NashConv in two-player Leduc.
\end{center}

\begin{center}
\begin{tabular}{|l|c|c|c|}
\hline
Parameter     & Value Set & Final Value \\
\hline
RL learning rate              & 1e-3, 1e-4, 1e-5 & 1e-4   \\
SL learning rate              & 1e-3, 1e-4, 1e-5 & 1e-5   \\
Anticipatory parameter        & 0.1, 0.5 & 0.5    \\
Reservoir buffer size         & 100000, 1000000 & 1000000 \\
\hline
\end{tabular}\\
Best values for exploitation against random bots in two-player Leduc.
\end{center}

\begin{center}
\begin{tabular}{|l|c|c|c|}
\hline
Parameter     & Value Set & Final Value \\
\hline
RL learning rate              & 1e-3, 1e-4, 1e-5 & 1e-3   \\
SL learning rate              & 1e-3, 1e-4, 1e-5 & 1e-4   \\
Anticipatory parameter        & 0.1, 0.5 & 0.5    \\
Reservoir buffer size         & 100000, 1000000 & 100000 \\
\hline
\end{tabular}\\
Best values for exploitation against cfr500 bots in two-player Leduc.
\end{center}

\subsubsection{NFSP parameters in Three-Player Leduc}

\begin{center}
\begin{tabular}{|l|c|c|c|}
\hline
Parameter     & Value Set & Final Value \\
\hline
RL learning rate              & 1e-3, 1e-4, 1e-5 & 1e-4    \\
SL learning rate              & 1e-3, 1e-4, 1e-5 & 1e-4    \\
Anticipatory parameter        & 0.1, 0.5 & 0.5     \\
Reservoir buffer size         & 100000, 1000000 & 1000000  \\
\hline
\end{tabular}\\
Best values for NashConv in three-player Leduc.
\end{center}

\begin{center}
\begin{tabular}{|l|c|c|c|}
\hline
Parameter     & Value Set & Final Value \\
\hline
RL learning rate              & 1e-3, 1e-4, 1e-5 & 1e-3    \\
SL learning rate              & 1e-3, 1e-4, 1e-5 & 1e-5    \\
Anticipatory parameter        & 0.1, 0.5 & 0.5     \\
Reservoir buffer size         & 100000, 1000000 & 100000  \\
\hline
\end{tabular}\\
Best values for exploitation versus random bots in three-player Leduc.
\end{center}

\begin{center}
\begin{tabular}{|l|c|c|c|}
\hline
Parameter     & Value Set & Final Value \\
\hline
RL learning rate              & 1e-3, 1e-4, 1e-5 & 1e-5    \\
SL learning rate              & 1e-3, 1e-4, 1e-5 & 1e-3    \\
Anticipatory parameter        & 0.1, 0.5 & 0.5     \\
Reservoir buffer size         & 100000, 1000000 & 100000  \\
\hline
\end{tabular}\\
Best values for exploitation versus cfr500 bots in three-player Leduc.
\end{center}

\subsection{Policy-Space Response Oracles (PSRO)}

In our Leduc Poker experiments, we use an alternating implementation that switches
periodically between computing the best response (oracle training phase) and meta-strategy learning phase where
the empirical payoff tensor is updated and meta-strategy computed.
We call this meta-game update frequency in the parameters below.
We also use $\Pi^T$ as described in
Section~\ref{sec:psro}, so the current set of policies includes (on the $k^{th}$ epoch) the currently training
oracle $\pi'$.
Finally, we add controlled exploration by linear annealing of the inverse temperature of the softmax policy head,
starting at 0 (uniform random) to 1.

Since the setting (action space, observation space, reward space, and network architecture) differ significantly
in the setting of Leduc poker, we try different values for the hyper-parameters. Our general methodology was to
manually try a few from subsets of sensible ranges on two-player Leduc, then use these values as starting points
for three-player. Finally the values in PSRO were used as starting points for DCH parameters.

For each parameter, we also give a rough sensitivity rating on a scale from 1 (not at all sensitive) to 5 (very sensitive)
Overall, we found that the algorithms were fairly robust to different parameter values in the range, and we note
some main general points: (i) their
values differed from those in the visual domains such as Atari and our gridworld games,
(ii) the most important parameter was the learning rate.

\subsubsection{PSRO Parameters in Two-Player Leduc Poker}


\begin{center}
\begin{tabular}{|l|c|c|c|}
\hline
Parameter     & Value Set & Final Value & SR \\
\hline
Learning rate              & 5e-5, 1e-4, 2.5e-4, 0.5e-4, 1e-3 & 5e-4 & 4 \\
Batch size                 & 4, 8, 16  & 16 & 1 \\
Replay buffer size         & 1e5, 5e5 & 5e5 & 1 \\
Sequence (unroll) length   & 2, 3, 8 & 2 & 2 \\
Entropy cost               & 0, 0.01, 0.1 & 0.1 & 2 \\
Exploration decay end      & 25000, 250000 & 250000 & 3\\
Target update period       & 1000, 5000, 10000 & 1000 & 2\\
Meta-game update frequency & 2500, 5000 & 2500 & 1\\
Episodes per epoch         & 25000 $\rightarrow$ \{ 5e4, 1e5, 2.5e5 \} & 25000 $\rightarrow$ 2.5e5 & 2\\
Trace parameter ($\lambda$) & 0.75, 0.9, 0.95, 1.0 & 0.95 & 1\\
Maximum epochs            & 20 & 20 & -- \\
\hline
\end{tabular}
\end{center}

SR stands for sensitivity rating. The exploration decay end is in number of steps taken by the ReActor oracle.

\subsubsection{PSRO Parameters in Three-Player Leduc Poker}


\begin{center}
\begin{tabular}{|l|c|c|c|}
\hline
Parameter     & Value Set & Final Value & SR \\
\hline
Learning rate              & 5e-5, 1e-4, 2.5e-4, 0.5e-4, 1e-3 & 5e-4 & 4 \\
Batch size                 & 4, 8, 16  & 16 & 1 \\
Replay buffer size         & 1e5, 5e5 & 5e5 & 1 \\
Sequence (unroll) length   & 2, 3, 8 & 2 & 2 \\
Entropy cost               & 0, 0.01, 0.1 & 0.1 & 2 \\
Exploration decay end      & 25000, 250000 & 250000 & 3\\
Target update period       & 1000, 5000, 10000 & 1000 & 2\\
Meta-game update frequency & 2500, 5000 & 2500 & 1\\
Episodes per epoch         & 25000 $\rightarrow$ \{ 5e4, 1e5, 2.5e5 \} & 25000 $\rightarrow$ 2.5e5 & 2\\
Trace parameter ($\lambda$) & 0.75, 0.9, 0.95, 1.0 & 1.0$^*$ & 1\\
Maximum epochs            & 20 & 20 & -- \\
\hline
\end{tabular}
\end{center}

$^*$The only value that changed moving to three-player Leduc was the trace parameter, $\lambda$, and it had
such a small effect that in the full runs we left $\lambda = 0.95$ for consistency.

\subsection{DCH}

\subsubsection{DCH Parameters in First-Person Gridworld Coordination Games}

In the first-person gridworld games, we use hyper-parameter settings that that are quite similar to the
default ReActor values~\cite{Gruslys17}, as noted above in Appendix~\ref{sec:alg_training_details}. In this
environment, we found that parameter values had little to no effect on the outcomes.

\subsubsection{DCH Parameters in Two-Player Leduc Poker}

%

In Leduc poker, there is one new parameter: the policy and meta-strategy save \& load frequencies
($T_{ms}$ in Algorithm~\ref{alg:dch}).
This is asynchronous analogue to the meta-game update frequency in PSRO.
The basic parameter tuning was more difficult for DCH due to large number of resources necessary.
Since we want to measure the tension between scalability and accuracy, we tune our hyper parameters
only on one value (1000) and include a sweep over $T_{ms} \in \{ 1000, 2500, 5000 \}$ in the full runs.
We also try the smaller value since the decoupled meta-solvers are online
and require more recent up-to-date estimates of the values.

\begin{center}
\begin{tabular}{|l|c|c|c|}
\hline
Parameter     & Value Set & Final Value & SR \\
\hline
Learning rate              & 1e-5, 5e-5, 1e-4, 2.5e-4, 0.5e-4, 1e-3 & 1e-4 & 4 \\
Batch size                 & 16 & 16 & -- \\
Replay buffer size         & 1e5, 5e5 & 5e5 & 1 \\
Sequence (unroll) length   & 2, 3, 8 & 8 & 2 \\
Entropy cost               & 0.1 & 0.1 & -- \\
Exploration decay end      & 25000, 250000 & 250000 & 3\\
Target update period       & 1000, 5000, 10000 & 10000 & 2\\
Trace parameter ($\lambda$) & 0.75, 0.9, 0.95, 1.0 & 0.95 & 1\\
Policy + meta-strategy update frequency & 1000, 2500, 5000 & 2500 & 1\\
\hline
\end{tabular}
\end{center}

\subsubsection{DCH Parameters in Three-Player Leduc Poker}

%

\begin{center}
\begin{tabular}{|l|c|c|c|}
\hline
Parameter     & Value Set & Final Value & SR \\
\hline
Learning rate              & 1e-5, 5e-5, 1e-4, 2.5e-4, 0.5e-4, 1e-3 & 5e-4 & 4 \\
Batch size                 & 16 & 16 & -- \\
Replay buffer size         & 1e5, 5e5 & 5e5 & 1 \\
Sequence (unroll) length   & 2, 3, 8 & 8 & 2 \\
Entropy cost               & 0.1 & 0.1 & -- \\
Exploration decay end      & 25000, 250000 & 250000 & 3\\
Target update period       & 1000, 5000, 10000 & 10000 & 2\\
Trace parameter ($\lambda$) & 0.75, 0.9, 0.95, 1.0 & 0.95 & 1\\
Policy + meta-strategy update frequency & 1000, 2500, 5000 & 2500 & 1\\
\hline
\end{tabular}
\end{center}

\subsection{Meta-Solvers}

For projected replicator dynamics, the average strategy value was tracked using a running average of 50 values.
The value of each policy was tracked using a running average of 10 values. The step size was set to
$\delta = 0.01$.


\section{Environments \label{sec:environments}}

\begin{figure}[h!]
\centering
\includegraphics[width=0.4\textwidth]{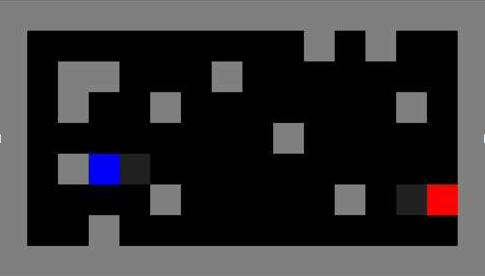}
\caption{Global view of Laser Tag small3 map (not seen by agents).
Agents see a localized view. Input to their networks is raw RGB pixels.}
\end{figure}

\begin{figure*}
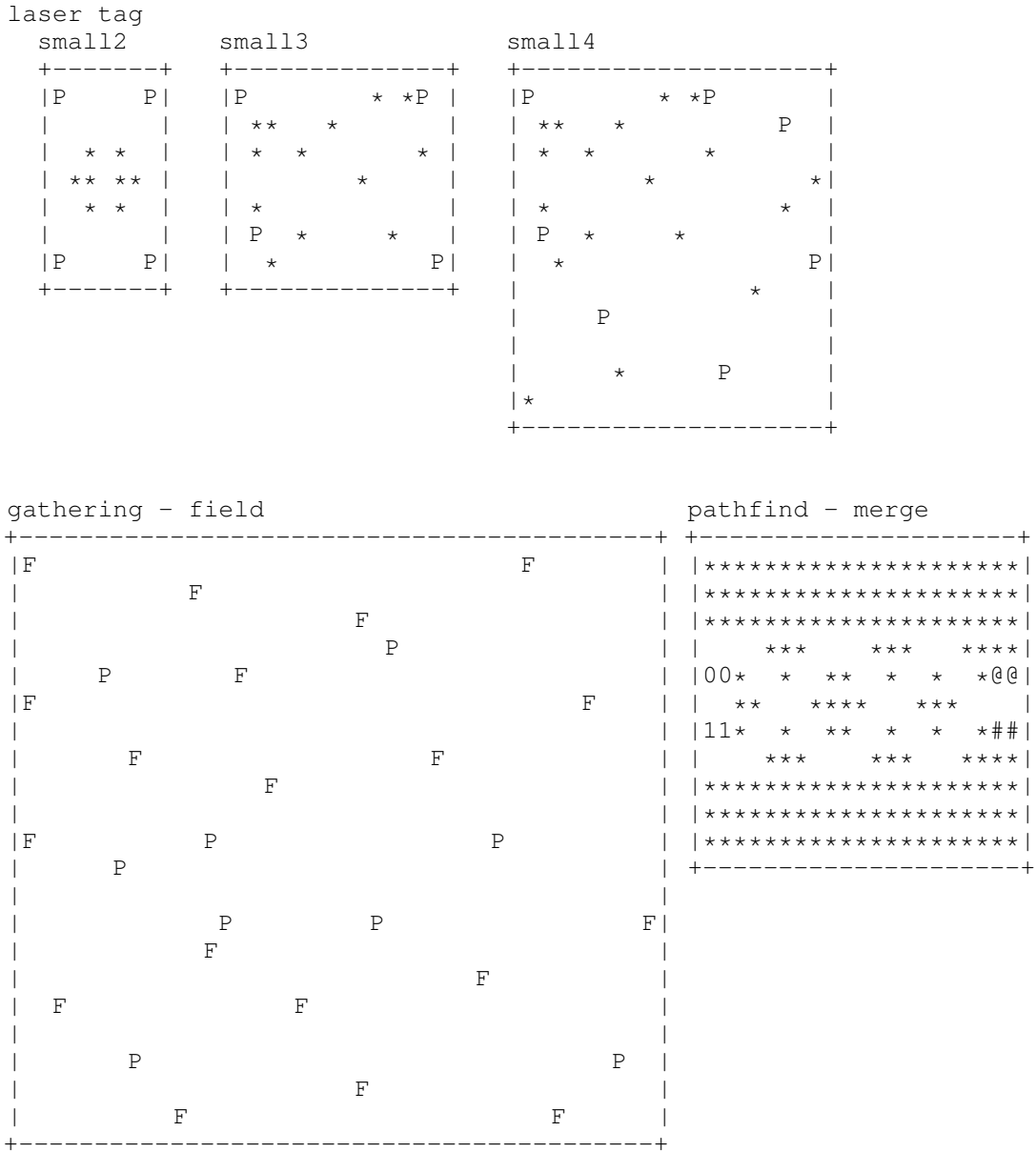

\begin{verbatim}
laser tag
  small2      small3             small4
  +-------+   +--------------+   +--------------------+
  |P     P|   |P        * *P |   |P        * *P       |
  |       |   | **   *       |   | **   *          P  |
  |  * *  |   | *  *       * |   | *  *       *       |
  | ** ** |   |        *     |   |        *          *|
  |  * *  |   | *            |   | *               *  |
  |       |   | P  *     *   |   | P  *     *         |
  |P     P|   |  *          P|   |  *                P|
  +-------+   +--------------+   |               *    |
                                 |     P              |
                                 |                    |
                                 |      *      P      |
                                 |*                   |
                                 +--------------------+


gathering - field                            pathfind - merge
+------------------------------------------+ +---------------------+
|F                                F        | |*********************|
|           F                              | |*********************|
|                      F                   | |*********************|
|                        P                 | |    ***    ***   ****|
|     P        F                           | |00*  *  **  *  *  *@@|
|F                                    F    | |  **   ****   ***    |
|                                          | |11*  *  **  *  *  *##|
|       F                   F              | |    ***    ***   ****|
|                F                         | |*********************|
|                                          | |*********************|
|F           P                  P          | |*********************|
|      P                                   | +---------------------+
|                                          |
|             P         P                 F|
|            F                             |
|                              F           |
|  F               F                       |
|                                          |
|       P                               P  |
|                      F                   |
|          F                        F      |
+------------------------------------------+
\end{verbatim}
\caption{Maps used in gridworld games. Here, {\tt P} is a spawn point for both players, {\tt *} are blocked/wall cells,
{\tt F} is food (apples), {\tt 0} and {\tt 1} are spawn points for
player 1 and 2 respectively, {\tt @} and {\tt \#}
are goal locations for player 1 and 2 respectively. \label{fig:maps}}
\end{figure*}

The maps we used are shown in Figure~\ref{fig:maps}.

\subsection{Representing Turn-based Games and Handling Illegal Moves \label{sec:games_repr}}

In poker, the number of legal actions that an agent can use
is a subset of the total number of unique actions in the entire game.
Also, the game is turn-based so only one player acts at a given time.

A modification to the standard environment is made so that policies can be defined over a
fixed set of actions and independent of the specific underlying RL algorithm
(i.e. DQN, ReAactor, etc.): the environment presents all 3 actions at all times; if an illegal
action is taken, then the agent receives a reward equal to the lower bound of the payoff
at a terminal node minus 1 ($= -14$ in Leduc), and a random legal move is chosen instead.
The resulting game is more complex and the agent
must first learn the rules of the game in addition to the strategy.
This also makes the game general-sum, so
CFR and exploitability were instead run on the original game. Exploitability of PSRO, DCH,
and NFSP policies are computed by first transforming the policies to legal ones by masking
out illegal moves and renormalizing.

Some RL algorithms process experience using transition tuples of the form \eg
$(s, a, r, s')$, or longer chains such as in ReActor. However, in turn-based games, given a trajectories
\[(s_t, a_t, r_t, s_{t+1}, a_{t+1}, r_{t+1}, s_{t+2}, \cdots),\]
the next state may not belong to the same player, so we construct player-specific tuples
of the form $(s_t, a_t, r_t, s_{t+1}, \cdots, s_{t+k}, \cdots)$, where $k$ is the number of steps
until it becomes the same player's turn again, so \eg in strictly-alternating games $k=2$.
Special cases are needed for terminal states, where all
players see the terminal state as the final transition.

\section{Results \label{sec:app_results}}

\begin{figure}
  \centering
    \includegraphics[width=\textwidth]{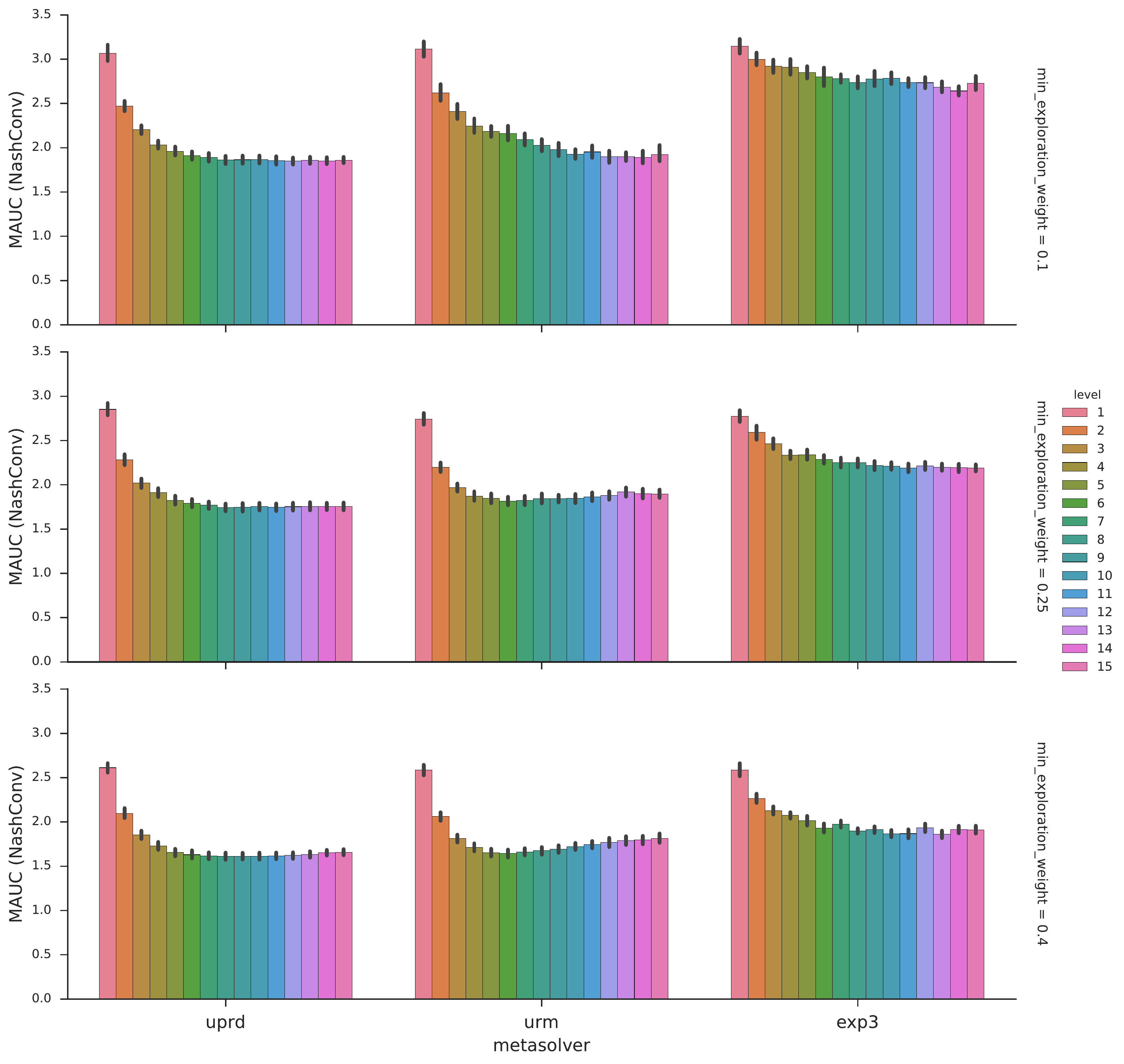}
    \caption{Effect of DCH parameters in two-player Leduc Poker (NashConv).}
    \label{fig:dhc_params_2pleduc_expl}
\end{figure}

\begin{figure}
  \centering
    \includegraphics[width=\textwidth]{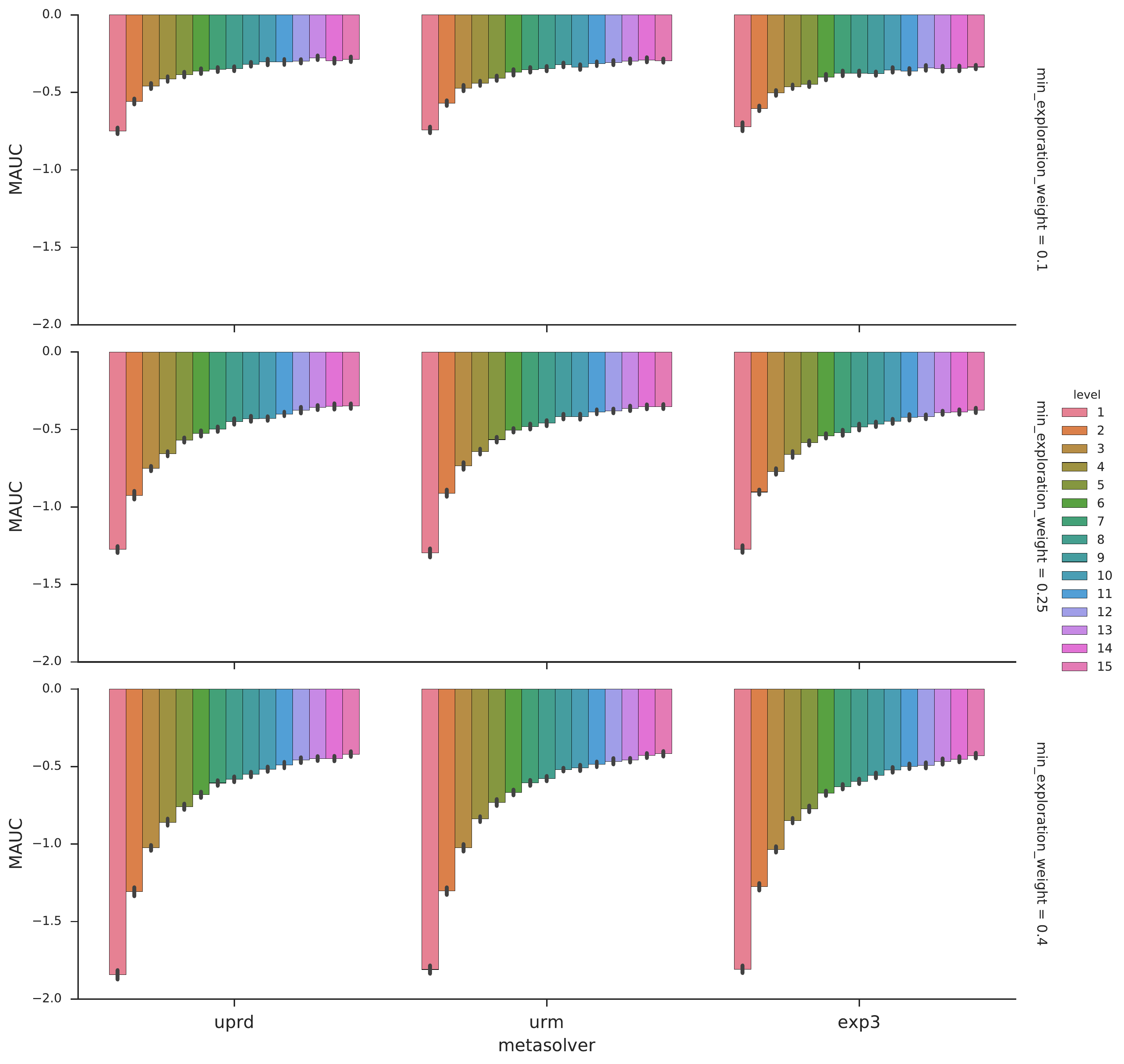}
    \caption{Effect of DCH parameters in two-player Leduc Poker (exploitation).}
    \label{fig:dhc_params_2pleduc_exploit}
\end{figure}

\begin{figure}
  \centering
    \includegraphics[width=0.75\textwidth]{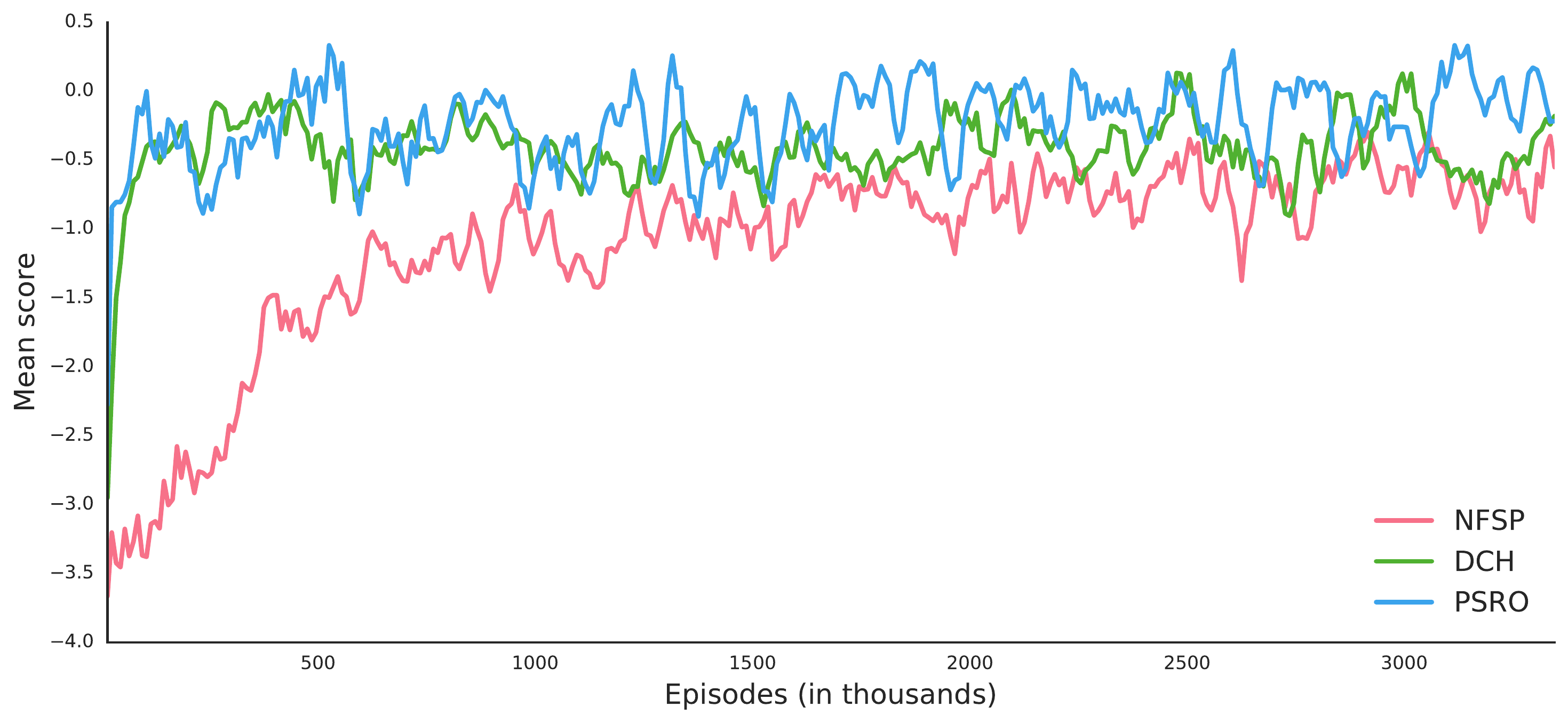}
    \caption{2-player: Exploitation vs. CFR500pure}
    \label{fig:exploit_2p_cfr500pure}
\end{figure}

\begin{figure}
  \centering
    \includegraphics[width=0.75\textwidth]{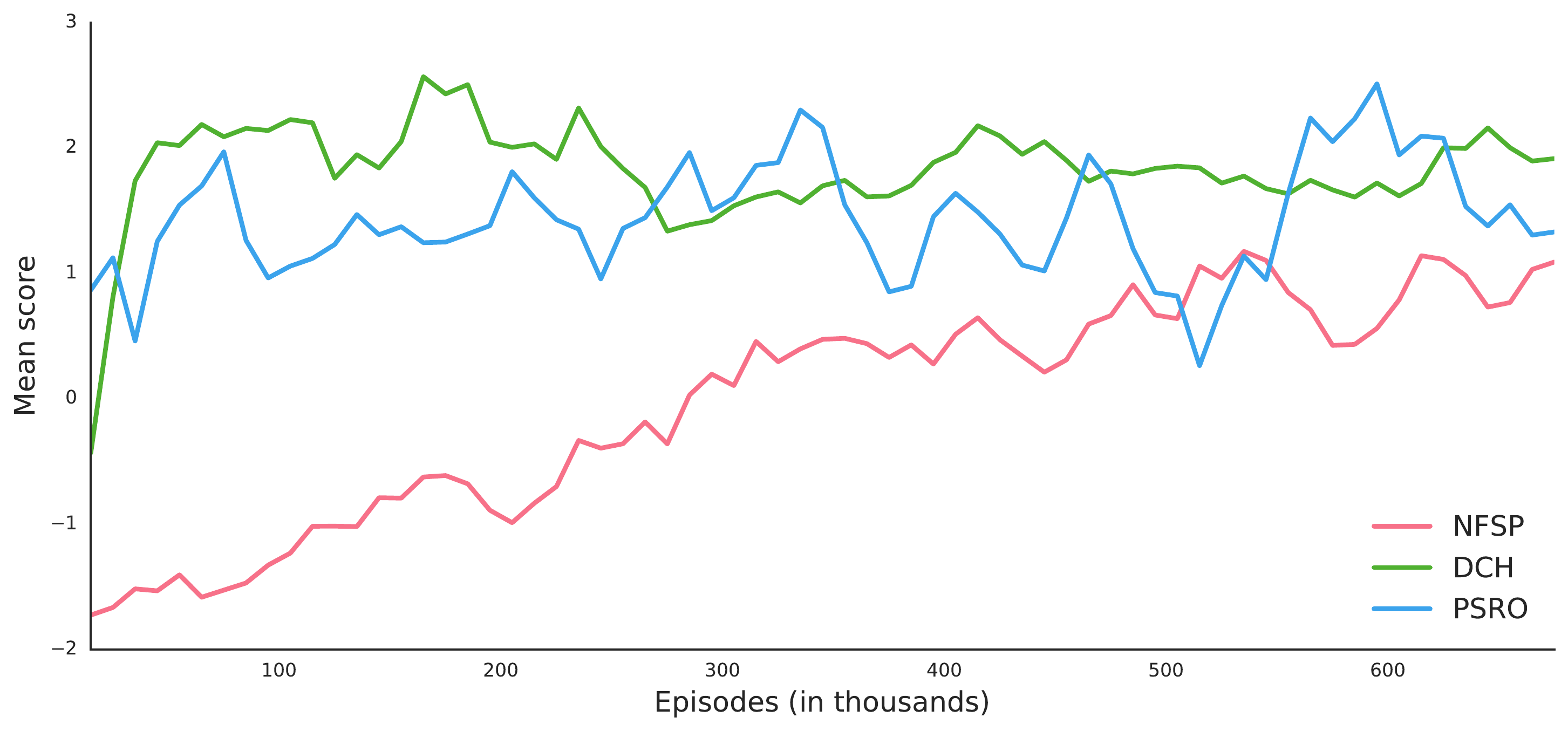}
    \caption{3-player: exploitation vs. random bots}
    \label{fig:exploit_3p_vs_random}
\end{figure}

\begin{figure}
  \centering
    \includegraphics[width=0.75\textwidth]{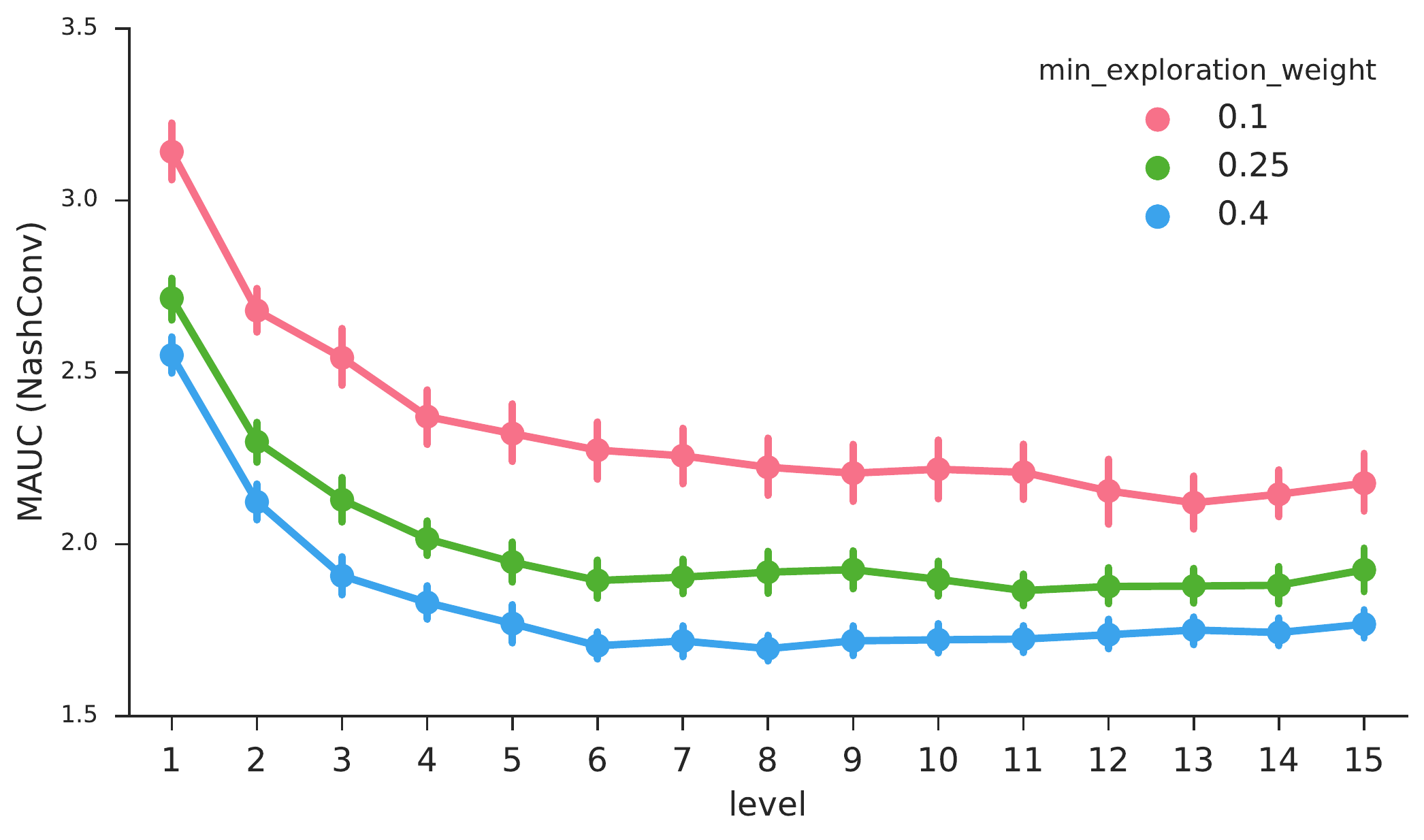}
    \caption{Effect of DCH parameter value of $\gamma$ on NashConv overall in two-player Leduc}
    \label{fig:min_expl_weight_1line}
\end{figure}

\begin{figure}
  \centering
    \includegraphics[width=0.75\textwidth]{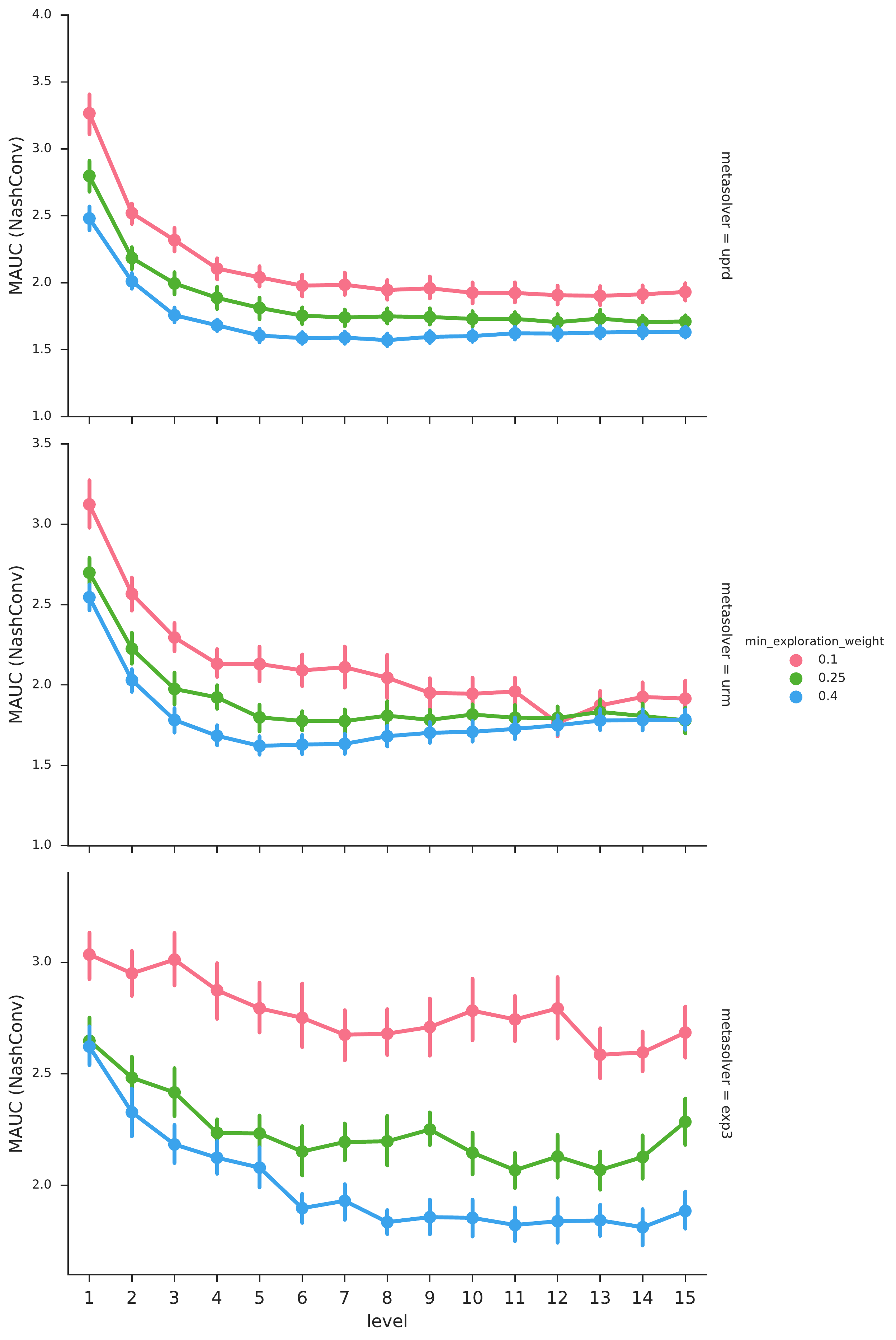}
    \caption{Effect of DCH parameter value of $\gamma$ on NashConv per meta-solver two-player Leduc}
    \label{fig:min_expl_weight_3line}
\end{figure}

See the following figures:
\begin{itemize}
\item Effect of DCH parameter values on NashConv in two-player Leduc: Figure~\ref{fig:dhc_params_2pleduc_expl}.
\item Effect of DCH parameter value of $\gamma$ on NashConv overall in two-player Leduc: Figure~\ref{fig:min_expl_weight_1line}.
\item Effect of DCH parameter value of $\gamma$ on NashConv per meta-solver two-player Leduc: Figure~\ref{fig:min_expl_weight_3line}.
\item Effect of DCH parameter values on exploitation in two-player Leduc: Figure~\ref{fig:dhc_params_2pleduc_exploit}.
\item Exploitation versus cfr500pure in two-player Leduc: Figure~\ref{fig:exploit_2p_cfr500pure}.
\item Exploitation versus random bots in three-player Leduc: Figure~\ref{fig:exploit_3p_vs_random}.
\end{itemize}

\subsection{Linear Regression Analysis to Interpret the Effect on DCH Parameters}

Using the data from the final runs, we also tested the effect of removing individual parameter settings
(value of $\gamma$, the meta-solver, levels, number of levels, meta-strategy update period, learning
rate) on the outcomes of exploitability and explotation in DCH.

We do this by fitting an ordinary least squares model to predict the
(exploitability or exploitation) value based on the parameter values,
via the Statsmodels Python module~\cite{seabold2010statsmodels}.

Below, we show the verbatim output of this analysis in Figure~\ref{fig:dch_param_regression}.
What this shows is the effect when removing individual settings of parameter values on the overall
prediction that includes all of the data. It does not (necessarily) show the best value for each parameter,
since the parameter values could combine in some complex non-linear way.

\begin{figure*}
{\scriptsize
\begin{verbatim}
                      DCH in 2-player Leduc for NashConv
                            OLS Regression Results
==============================================================================
Dep. Variable:                    AUC   R-squared:                       0.555
Model:                            OLS   Adj. R-squared:                  0.555
Method:                 Least Squares   F-statistic:                     897.2
Date:                Mon, 06 Nov 2017   Prob (F-statistic):               0.00
Time:                        16:18:25   Log-Likelihood:                -1318.8
No. Observations:                6480   AIC:                             2658.
Df Residuals:                    6470   BIC:                             2725.
Df Model:                           9
Covariance Type:            nonrobust
=====================================================================================================
                                        coef    std err          t      P>|t|      [0.025      0.975]
-----------------------------------------------------------------------------------------------------
Intercept                             3.1009      0.015    209.735      0.000       3.072       3.130
C(learning_rate)[T.0.0001]           -0.0551      0.007     -7.472      0.000      -0.070      -0.041
C(min_exploration_weight)[T.0.25]    -0.3261      0.009    -36.109      0.000      -0.344      -0.308
C(min_exploration_weight)[T.0.40]    -0.4791      0.009    -53.049      0.000      -0.497      -0.461
C(metasolver)[T.uprd]                -0.3974      0.009    -44.000      0.000      -0.415      -0.380
C(metasolver)[T.urm]                 -0.3359      0.009    -37.195      0.000      -0.354      -0.318
C(load_period)[T.2500]                0.0328      0.009      3.630      0.000       0.015       0.050
C(load_period)[T.5000]                0.0998      0.009     11.047      0.000       0.082       0.117
meta_strategy_update_period           0.0004      0.001      0.384      0.701      -0.002       0.003
np.log(level)                        -0.2542      0.005    -52.075      0.000      -0.264      -0.245
==============================================================================
Omnibus:                      186.685   Durbin-Watson:                   0.982
Prob(Omnibus):                  0.000   Jarque-Bera (JB):              223.211
Skew:                           0.366   Prob(JB):                     3.39e-49
Kurtosis:                       3.539   Cond. No.                         28.3
==============================================================================


                      DCH in 3-player Leduc for NashConv
                            OLS Regression Results
==============================================================================
Dep. Variable:                    AUC   R-squared:                       0.750
Model:                            OLS   Adj. R-squared:                  0.750
Method:                 Least Squares   F-statistic:                     3239.
Date:                Mon, 06 Nov 2017   Prob (F-statistic):               0.00
Time:                        16:27:45   Log-Likelihood:                -4302.7
No. Observations:                9720   AIC:                             8625.
Df Residuals:                    9710   BIC:                             8697.
Df Model:                           9
Covariance Type:            nonrobust
======================================================================================================
                                         coef    std err          t      P>|t|      [0.025      0.975]
------------------------------------------------------------------------------------------------------
Intercept                              5.7709      0.015    376.465      0.000       5.741       5.801
C(learning_rate)[T.0.0001]             0.2603      0.008     34.046      0.000       0.245       0.275
C(min_exploration_weight)[T.0.25]     -0.4380      0.009    -46.773      0.000      -0.456      -0.420
C(min_exploration_weight)[T.0.40]     -0.6584      0.009    -70.311      0.000      -0.677      -0.640
C(metasolver)[T.uprd]                 -0.9475      0.009   -101.178      0.000      -0.966      -0.929
C(metasolver)[T.urm]                  -0.9550      0.009   -101.984      0.000      -0.973      -0.937
C(load_period)[T.2500]                -0.0231      0.009     -2.464      0.014      -0.041      -0.005
C(load_period)[T.5000]                 0.0757      0.009      8.079      0.000       0.057       0.094
meta_strategy_update_period           -0.0012      0.001     -1.056      0.291      -0.003       0.001
np.log(level)                         -0.4797      0.005    -94.801      0.000      -0.490      -0.470
==============================================================================
Omnibus:                     2064.928   Durbin-Watson:                   0.916
Prob(Omnibus):                  0.000   Jarque-Bera (JB):             7655.088
Skew:                           1.030   Prob(JB):                         0.00
Kurtosis:                       6.829   Cond. No.                         28.3
==============================================================================
\end{verbatim}
}
\caption{Output of regression tests to interpret effects of DCH parameters. For these results, we used
a MAUC of the NashConv over the most recent 128 values.\label{fig:dch_param_regression}}
\end{figure*}

We observe one particularly interesting point from this analysis: both the meta-solver and the level
structure in DCH seem to have a
stronger effect on lowering NashConv in three-player Leduc than in two-player Leduc.
This could be due to the fact that the two-player game is smaller, and the DCH is struggling to find a precise
Nash equilibrium since it is instead maximizing reward against the specific subset of (15) oracles.
However, it could also be because opponent modeling and learning to anticipate actions from the other players
is more important when learning to play games with more than two players. We hope to investigate this further:
particularly the link between training regimes for multiagent reinforcement learning that
produce policies capable of generalizing to arbitrary behavior online (during execution), and whether and how
this could lead to an implicitly-encoded Theory of Mind.

\subsection{CFR Exploitability in Leduc \label{sec:expl_cfr}}

\begin{figure}[h!]
  \centering
    \includegraphics[width=0.75\textwidth]{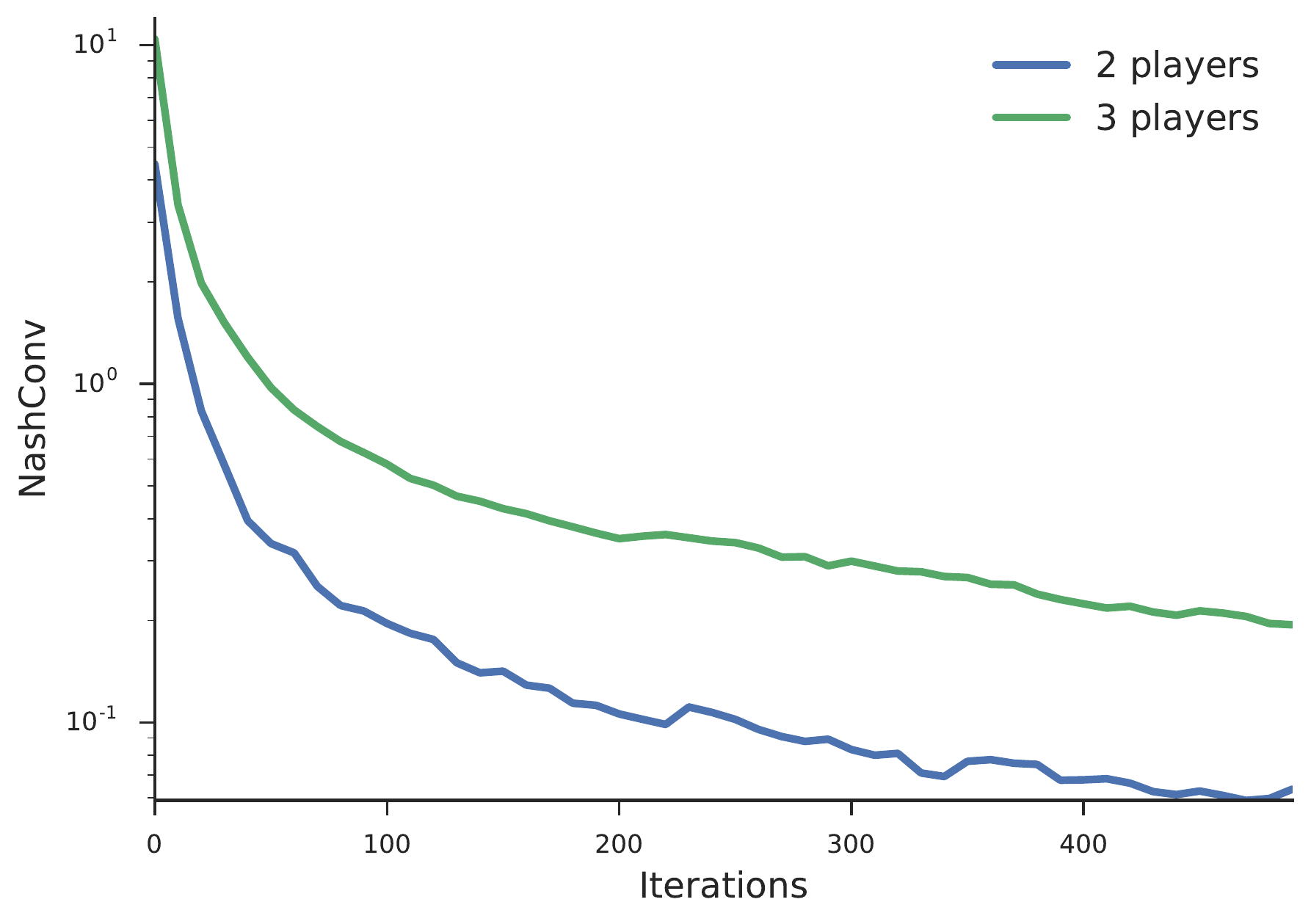}
    \caption{NashConv values of vanilla CFR in two-player and three-player Leduc Poker.}
    \label{fig:expl_cfr_2pleduc}
\end{figure}

The convergence graph of vanilla CFR is shown in Figure~\ref{fig:expl_cfr_2pleduc}.
No abstractions were used.

The \textsc{NashConv} values reach at iteration 500 are $0.063591$ for two-player,
and $0.194337$ for three-player.

The value (\ie expected winnings for first player under any exact Nash equilibrium)
of two-player Leduc is $-0.085606424078$, so the second player has a slight advantage. To compute this
number, the exact Nash equilibrium was obtained using sequence-form
linear programming~\cite{SequenceFormLPs}; see also \cite[Section 5.2.3]{Shoham09}.

\subsection{Computing the Explicit Meta-Policy for Exploitability in PSRO/DCH}

In PSRO and DCH, the meta-strategy $\sigma_i$ is a distribution over policies $\Pi_i$.
The combination of the two encodes a single stochastic policy $\pi_i^{\sigma}$
that can be obtained
by doing a pass through the game tree. In terms of computational game theory, this
means applying Kuhn's theorem~\cite{Kuhn53} to convert a mixed strategy to a behavior strategy.

For each information state $I$, the probability of taking action $a$ is $\pi_i^{\sigma}(I, a)$.
This can be computed by computing weights $W_{I,a}$ which is the sum over global states
$s \in I$ of reach probabilities to get to $s$ under each policy $\pi_{i,k} \in \Pi_i$
times the probability $\sigma_i(\pi_{i,k}) \cdot \pi_i(s,a)$.
Then the final stochastic policy is obtained by
\[ \pi_i^{\sigma}(s,a) = \frac{W_{I,a}}{\sum_{a'} W_{I,a'}}. \]
(Note that opponents' policies need not be considered in the computation as they would cancel
when $W_{I,a}$ is normalized.)

\end{document}